\documentclass[]{article}

\PassOptionsToPackage{numbers}{natbib}

\usepackage[final]{neurips_2022}

\usepackage[utf8]{inputenc} 
\usepackage[T1]{fontenc}    
\usepackage{hyperref}       
\usepackage{url}            
\usepackage{booktabs}       
\usepackage{amsfonts}       
\usepackage{nicefrac}       
\usepackage{microtype}      
\usepackage{xcolor}         
\usepackage{amsmath, amssymb, amsthm, mathtools, bm}
\usepackage{enumitem}
\usepackage{float}
\usepackage{caption}
\usepackage[justification=centering]{subcaption}
\usepackage{tikz}
\usetikzlibrary{arrows.meta,calc,patterns,angles,quotes,bending}
\usepackage{pgfplots}
\pgfplotsset{compat=newest}
\usepackage{kotex}
\usepackage{booktabs, makecell, multirow, diagbox}
\usepackage{adjustbox}
\usepackage{derivative}
\usepackage{textcomp}
\usepackage{graphicx}
\usepackage{wrapfig}
\usepackage{tabularx}
\usepackage{babel}
\usepackage{mwe}
\usepackage{stackengine}
\usepackage{varwidth}
\usepackage{fontawesome}
\usepackage{algorithm, algpseudocode}

\newsavebox\tmpbox
\newtheorem{lemma}{Lemma}
\newtheorem{thm}{Theorem}
\newtheorem{defn}{Definition}
\newtheorem{example}[lemma]{Example}

\newcommand{\norm}[1]{\left\lVert#1\right\rVert}

\title{Invertible~Monotone~Operators~for~Normalizing~Flows}

\author{%
  Byeongkeun Ahn$^{1}$, Chiyoon Kim$^{1}$, Youngjoon Hong$^{2\ast}$, Hyunwoo J.~Kim$^{1}\thanks{Corresponding authors.}$ \\
  Korea University$^{1}$, 
  Sungkyunkwan University$^{2}$ \\
  \texttt{\{byeongkeunahn, kimchiyoon, hyunwoojkim\}@korea.ac.kr} \\
  \texttt{hongyj@skku.edu} \\
}

\def\eg{\emph{e.g}.}

\begin{document}

\maketitle

\begin{abstract}
    Normalizing flows model probability distributions by learning invertible transformations that transfer a simple distribution into complex distributions. Since the architecture of ResNet-based normalizing flows is more flexible than that of coupling-based models, ResNet-based normalizing flows have been widely studied in recent years.
Despite their architectural flexibility, it is well-known that the current ResNet-based models suffer from constrained Lipschitz constants. In this paper, we propose the \emph{monotone formulation} to overcome the issue of the Lipschitz constants using monotone operators and provide an in-depth theoretical analysis. Furthermore, we construct an activation function called Concatenated Pila (CPila) to improve gradient flow. The resulting model, \emph{Monotone Flows}, exhibits an excellent performance on multiple density estimation benchmarks (MNIST, CIFAR-10, ImageNet32, ImageNet64). Code is available at \href{https://github.com/mlvlab/MonotoneFlows}{\texttt{https://github.com/mlvlab/MonotoneFlows}}.

\end{abstract}

\section{Introduction}
    \label{sec:intro}

Normalizing flows~\cite{rezende2015variational, Nice} are a method for constructing complex distributions by transforming a probability density through a series of invertible transformations. Normalizing flows are trained using a plain log-likelihood function, and they are capable of exact density evaluation and efficient sampling. Applications include image generation~\cite{Nice, RealNVP, GLOW, i-ResNets, ResidualFlows, i-DenseNets, CPFlow}, image super-resolution~\cite{SRFlow, SRFlow-DA, adFlow}, image noise modelling~\cite{NoiseFlow, InvDN}, audio synthesis~\cite{Glow-TTS, WaveGlow}, anomaly detection~\cite{CS-Flow, CFLOW-AD}, and computational physics~\cite{albergo2019flow, gao2020event, nachman2020anomaly, SmoothNF}, highlighting the importance of developing expressive normalizing flows.

However, a major hurdle to designing normalizing flow architectures is that not only should each transformation be invertible but also its Jacobian determinant needs to be tractable~\cite{Kobyzev2021NormalizingFA, Papamakarios2021NormalizingFF}. Despite multiple efforts to satisfy the aforementioned constraints~\cite{Nice, RealNVP, GLOW, ConvExp, Woodbury}, the expressive power of the normalizing flow transformations stays limited due to the adoption of specialized architectures~\cite{i-ResNets}.

In this regard, ResNet-based normalizing flows~\cite{i-ResNets, ResidualFlows, i-DenseNets} are a compelling alternative since they do not impose any structural restrictions. However, they keep the Lipschitz constant of the residual branch less than $1$ in order to ensure invertibility. As a result, the Lipschitz constant of each residual block is less than $2$. This causes a serious issue on the expressive power since a large Lipschitz constant is often required in converting between distributions. A recent work~\cite{ImpFlows} tackled this problem, but as we show in this paper, {their} method is equivalent to a simple rescaling of ResNet-based normalizing flows, which has a similar expressive power.

In this work, we propose the approach of \emph{monotone operators} to alleviate the Lipschitz constraint. More precisely, in order to parameterize monotone operators, we use the well-known fact from monotone operator theory: {\it the Cayley operator of maximally monotone operators (MMOs) is a 1-Lipschitz function}. The residual branch of ResNet-based normalizing flows can be used to parameterize the Cayley operators.

In summary, our contributions are as follows:
\begin{itemize}
    \item We propose \emph{Monotone Flows}, which greatly loosens the Lipschitz constraint while retaining invertibility and architectural flexibility.
	\item We derive the \emph{monotone formulation}, a monotone operator-based normalizing flow by parametrizing the Cayley operator, and provide efficient training and inference schemes.
	\item We propose a new activation function called Pila and its variant Concatenated Pila (CPila) to alleviate the saturated gradient problem.
	\item We theoretically analyze the expressive power of the monotone formulation and related models \cite{ResidualFlows, ImpFlows}, and prove our formulation outperforms the other models in practice.
	\item {Monotone Flows consistently outperform comparable baseline normalizing flows on multiple image density estimation benchmarks as well as on 2D toy datasets.} In addition, ablation studies demonstrate the effectiveness of the proposed methods.
\end{itemize}

\section{Preliminaries}
\label{sec:prelim}

\vspace{-0.1cm}
The monotone formulation is a normalizing flow that parametrizes monotone operators in terms of Cayley operators. The parametrization, which must be 1-Lipschitz, and the methods for log determinant computation, are borrowed from ResNet-based normalizing flows. Also, the resulting model is an implicit neural network involving inverse functions. Here, we briefly review these topics.

\noindent\textbf{Normalizing flows} ~\cite{rezende2015variational} explicitly model probability distributions by transforming a tractable source distribution, \eg, Gaussian, into a target distribution with a differentiable invertible transform $F: \mathbb{R}^n\rightarrow\mathbb{R}^n$. Suppose $F$ maps a sample space $X$ to a latent space $Z$. Given a source distribution $p_Z(z)$, the change-of-variables theorem yields the log-likelihood of any $x\in X$:
\vspace{-0.1cm}
\begin{align}\label{eq:change_of_variables}
	\log p_X(x) = \log p_Z(z) + \log|\det J_F|,
	\vspace{-0.1cm}
\end{align}
where $z = F(x)$ and $J_F = \partial z/\partial x$ is the Jacobian. 
Normalizing flows are beneficial since the
evaluation of (\ref{eq:change_of_variables}) yields the exact density $p_X(x)$.
Furthermore, target data $x\sim p_X(x)$ can be efficiently sampled by first sampling $z \sim p_Z(z)$ and then mapping $z$ to $X$ by $x = F^{-1}(z)$. When $F$ consists of multiple transformations applied in succession, each transformation is called a \emph{flow}.

\noindent\textbf{ResNet-based normalizing flows} have been initially proposed in i-ResNets~\cite{i-ResNets}, which are built with a variant of the residual blocks $R(x) = x + G(x)$. By requiring $G(x)$ to be a contraction map, they ensured the invertibility of $R(x)$ with the Banach fixed point theorem. The contractiveness of $G(x)$ is implemented by spectral normalization~\cite{SpectralNorm} and 1-Lipschitz activation functions. However, they use a biased truncation of the log Taylor series for computing the log determinant. Residual Flows~\cite{ResidualFlows} address this issue with an unbiased estimator based on randomized truncation. They also introduce a new activation function LipSwish to avoid the gradient saturation problem. i-DenseNets~\cite{i-DenseNets} improve Residual Flows with a DenseNet variant accompanying learnable concatenation coefficients. They also introduce the Concatenated LipSwish activation function to further improve the gradient flow.

\noindent\textbf{Implicit neural networks} map an input $x$ to an output $y$ through an implicit equation $F(x,y) = 0$, in contrast with explicit models of the form $y = F(x)$. The main advantage of implicit mappings is their ability to express a richer family of functions under the same parameter budget \cite{DEQ}. Implicit Flows \cite{ImpFlows} propose to use invertible implicit neural networks in normalizing flows and study an example where residual blocks and the inverse of residual blocks are stacked in an alternating fashion.

\definecolor{darkgreen30}{rgb}{0,0.67,0}
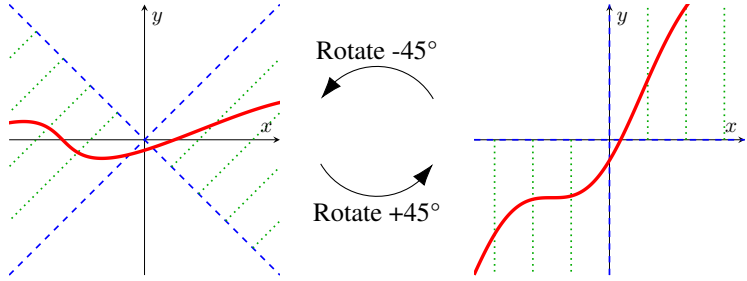
\begin{figure}[t]
	\centering
	\vspace{-0.7cm}
	\begin{tikzpicture}[
		declare function={
			f(\x) = (\x) + (sin(deg((\x)-1))) + (0.01*((\x)+2)*((\x)+2)*((\x)+2));
		},
		samples=1000,
		scale=0.8
		]
		\begin{axis}[
			axis x line=middle, axis y line=middle,
			ymin=-5, ymax=5, ylabel=$y$,y=0.45cm,
			xmin=-5, xmax=5, xlabel=$x$,x=0.45cm,
			ticks=none
			]
			\draw[-, blue, dashed, thick] (-5,5) -- (5,-5);
			\draw[-, blue, dashed, thick] (-5,-5) -- (5,5);
			\draw[-, darkgreen30, dotted, thick] (-4,4) -- (-5,3);
			\draw[-, darkgreen30, dotted, thick] (-3,3) -- (-5,1);
			\draw[-, darkgreen30, dotted, thick] (-2,2) -- (-5,-1);
			\draw[-, darkgreen30, dotted, thick] (-1,1) -- (-5,-3);
			\draw[-, darkgreen30, dotted, thick] (4,-4) -- (5,-3);
			\draw[-, darkgreen30, dotted, thick] (3,-3) -- (5,-1);
			\draw[-, darkgreen30, dotted, thick] (2,-2) -- (5,1);
			\draw[-, darkgreen30, dotted, thick] (1,-1) -- (5,3);
			\addplot[red, domain=-5:5, ultra thick, opacity=1.0]({(x+f(x))/sqrt(2)}, {(-x+f(x))/sqrt(2)});
		\end{axis}
	\end{tikzpicture}
	\begin{tikzpicture}[scale=0.72]
		\node (E) at (1.5,0) {};
		\node (W) at (-1.5,0) {};
		\node (S) at (0,-2.5) {};
		\node (N) at (0,2.5) {};
		\draw[-{Latex[length=3mm,width=2mm,flex=1.0]}] (210:1.2cm) arc (210:330:1.2cm);
		\node at (0,-1.5) {Rotate +45\textdegree};
		\draw[-{Latex[length=3mm,width=2mm,flex=1.0]}] (30:1.2cm) arc (30:150:1.2cm);
		\node at (0,1.5) {Rotate -45\textdegree};
	\end{tikzpicture}
	\begin{tikzpicture}[
		declare function={
			f(\x) = (\x) + (sin(deg((\x)-1))) + (0.01*((\x)+2)*((\x)+2)*((\x)+2));
		},
		samples=1000,
		scale=0.8
		]
		\begin{axis}[
			axis x line=middle, axis y line=middle,
			ymin=-5, ymax=5, ylabel=$y$,y=0.45cm,
			xmin=-5, xmax=5, xlabel=$x$,x=0.45cm,
			ticks=none
			]
			\draw[-, blue, dashed, thick] (-5,0) -- (5,0);
			\draw[-, blue, dashed, thick] (0,-5) -- (0,5);
			\draw[-, darkgreen30, dotted, thick] ({-3*sqrt(2)},0) -- ({-3*sqrt(2)},-5);
			\draw[-, darkgreen30, dotted, thick] ({-2*sqrt(2)},0) -- ({-2*sqrt(2)},-5);
			\draw[-, darkgreen30, dotted, thick] ({-1*sqrt(2)},0) -- ({-1*sqrt(2)},-5);
			\draw[-, darkgreen30, dotted, thick] ({3*sqrt(2)},0) -- ({3*sqrt(2)},5);
			\draw[-, darkgreen30, dotted, thick] ({2*sqrt(2)},0) -- ({2*sqrt(2)},5);
			\draw[-, darkgreen30, dotted, thick] ({1*sqrt(2)},0) -- ({1*sqrt(2)},5);
			\addplot[red, domain=-5:5, ultra thick, opacity=1.0]{f(x)};
		\end{axis}
	\end{tikzpicture}
	\caption{The duality between 1-Lipschitz operators (left) and monotone operators (right) in 1D.}
	\label{fig:duality_1d}
\end{figure}

\noindent\textbf{Monotone operator} $F:\mathbb{R}^n\rightarrow\mathbb{R}^n$ is an operator that satisfies $\langle u-v,x-y\rangle\ge0$ for all $x,y\in\mathbb{R}^n$ and all $u\in F(x)$, $v\in F(y)$. \footnote{We adopt the notation $F(x) = \lbrace y | (x,y)\in F\rbrace$ since $F$ is an operator. However, it suffices to consider single-valued functions defined at every point in our development of theory, in which case the definition reduces to $\langle F(x)-F(y),x-y\rangle\ge 0$.} This definition generalizes monotonic functions in $\mathbb{R}$, to $\mathbb{R}^n$. The \emph{resolvent} of a monotone operator $F$ is defined as $R_F = (\mathrm{Id}+F)^{-1}$, where $\mathrm{Id}$ is the identity operator. The \emph{Cayley operator} of $F$ is defined as $C_F = 2R_F - \mathrm{Id}$. It is well-known that the Cayley operators of monotone operators are exactly 1-Lipschitz operators. The Cayley operator can be interpreted as a 45-degrees rotation (up to constant scaling) of the corresponding monotone operator in $\mathbb{R}^n\times\mathbb{R}^n$; see Figure~\ref{fig:duality_1d} for a 1D illustration. Minty surjectivity theorem~(\cite{minty1962} and \cite[Theorem 21.1]{bauschke2011convex}) shows that the Cayley operators of \emph{maximally} monotone operators are exactly 1-Lipschitz \emph{functions}, which are defined at every point. We formally state this correspondence in Theorem~\ref{thm:duality}. See \cite{Ryu2016APO} for a survey and \cite{bauschke2011convex} for a comprehensive treatment on monotone operators.

\section{Monotone Flows}
    \vspace{-0.1cm}
In this section, we propose a novel approach, the \emph{monotone formulation}, and describe the training and inference procedure. Moreover, we introduce a new activation function Concatenated Pila (CPila).

\subsection{Motivation from monotone operator theory}
\vspace{-0.1cm}

ResNet-based normalizing flows consist of blocks of the form $R(x) = x + G(x)$, where $G(x)$ is a parametrized neural network. If $G(x)$ is a contraction (the Lipschitz constant $\mathrm{Lip}(G) < 1$), the Banach fixed point theorem shows that $R(x)$ is invertible. In this case, $\mathrm{Lip}(R)$ is strictly less than $2$, resulting in a severe limitation on the expressive power of the neural network $R$ since a large Lipschitz constant is required in general to transform the source distribution into the target distribution.

However, as shown in the next example, $G(x)$ need not be a contraction for $R(x)$ to be invertible.
\begin{example}\label{ex:invertibleGxNotContractive}
	Consider $G_1(x) = 5x$. Then, $R_1(x) = x + G_1(x) = 6x$ is invertible even though $G_1(x)$ is not a contraction.
\end{example}
\vspace{-0.2cm}
Hence, when considering a relaxed condition and parametrization, it is possible to enhance the expressive power to include a larger class of functions as in Example~\ref{ex:invertibleGxNotContractive}. For this, we introduce monotone operators which have been widely studied in the fields of functional analysis and partial differential equations. We first recall various definitions of the monotonicity of operators in $\mathbb{R}^n$:
\begin{defn}
	Consider an operator $F: \mathbb{R}^n\rightarrow\mathbb{R}^n$.
	\begin{enumerate}[label=(\roman*), leftmargin=1.0cm]
		\vspace{-0.2cm}
		\item $F$ is \emph{monotone} if for all $x,y\in\mathbb{R}^n$, $\langle F(x)-F(y),x-y\rangle \ge 0$.
		\vspace{-0.2cm}
		\item $F$ is \emph{strictly monotone} if for all $x,y\in\mathbb{R}^n$ $(x\neq y)$, $\langle F(x)-F(y),x-y\rangle > 0$.
		\vspace{-0.2cm}
		\item $F$ is \emph{$\eta$-strongly monotone} $(\eta>0)$ if for all $x, y\in\mathbb{R}^n$, $\langle F(x)-F(y),x-y\rangle \ge \eta \lVert x-y\rVert_2^2$.
		\vspace{-0.2cm}
		\item $F$ is \emph{maximally monotone} if $F$ is monotone and not a proper subset of any monotone operator.
	\end{enumerate}
	\vspace{-0.3cm}
\end{defn}
It is noteworthy that monotonicity does not guarantee injectivity nor surjectivity, and strict monotonicity guarantees injectivity but not surjectivity. However, we have the following theorem:
\begin{thm}\label{thm:stronglyMonotone}
	An $\eta$-strongly monotone continuous function $F: \mathbb{R}^n\rightarrow\mathbb{R}^n$ is invertible for any $\eta>0$.
    \vspace{-0.2cm}
\end{thm}
\textit{Proof.} See Corollary 20.28 and Proposition 22.11 in~\cite{bauschke2011convex}.

Theorem~\ref{thm:stronglyMonotone} easily shows that $R_1(x)$ in Example~\ref{ex:invertibleGxNotContractive} is invertible. 
Moreover, if $G(x)$ is $L$-Lipschitz $(L < 1)$, $R(x)$ is $(1-L)$-strongly monotone (see Appendix~\ref{apn:residual_formulation_strongly_monotone}) and thus invertible by Theorem~\ref{thm:stronglyMonotone}.

As in Example~\ref{ex:invertibleGxNotContractive}, the ResNet-based formulation of normalizing flows does not include all possible invertible and monotone functions. To address this issue, we directly parameterize monotone operators in terms of their Cayley operators. We first formally state the rationale behind our construction.
\begin{thm}\label{thm:duality}
    Let $F: \mathbb{R}^n\rightarrow\mathbb{R}^n$ be an operator and $C_F = 2(\mathrm{Id}+F)^{-1} - \mathrm{Id}$ be its Cayley operator. Then the followings hold.
    \begin{enumerate}[label=(\roman*), leftmargin=1.0cm]
		\vspace{-0.2cm}
        \item $F$ is monotone $\Leftrightarrow$ $C_F$ is 1-Lipschitz.
		\vspace{-0.2cm}
        \item $F$ is maximally monotone $\Leftrightarrow$ $C_F$ is 1-Lipschitz and $\mathrm{dom}\,C_F=\mathbb{R}^n$.
    \end{enumerate}
    \vspace{-0.2cm}
\end{thm}
\textit{Proof.} See Proposition 23.8 and Proposition 4.4 in~\cite{bauschke2011convex}.

Theorem~\ref{thm:duality} shows that each 1-Lipschitz function characterizes a unique MMO since the Lipschitz continuity of $C_F$ implies $C_F$ is single-valued. Hence, the Lipschitz continuous parametrizations of the residual branch in ResNet-based models fit naturally for specifying $C_F$.

We now revisit Example~\ref{ex:invertibleGxNotContractive} and apply Theorem~\ref{thm:duality}. 
The Cayley operator of $R_1(x) = 6x$ is given by
$C_{R_1}(x) = 2(\mathrm{Id}+R_1)^{-1}(x) - x = -(5/7)x$, which is indeed 1-Lipschitz in accord with Theorem~\ref{thm:duality}.

In general, however, MMOs can be multivalued or undefined at some points. However, when the Cayley operator $C_F$ is an $L$-Lipschitz function with $L<1$, then $F$ is $(1-L)/(1+L)$-strongly monotone and $(1+L)/(1-L)$-Lipschitz continuous; see Appendix~\ref{sec:monotone_op_L_Lipschitz_Cayley} for a complete derivation. 
Theorem~\ref{thm:stronglyMonotone} then implies $F$ is invertible. 
Hence, we use $L$-Lipschitz functions for Cayley operators.

\subsection{Description of the monotone formulation}

\begin{defn}\label{defn:monotone_flow}
	Let $G: \mathbb{R}^n\rightarrow\mathbb{R}^n$ be a {Lipschitz-continuous} function with Lipschitz constant $L < 1$. The monotone formulation of $G$ is defined as the following function $F: \mathbb{R}^n\rightarrow\mathbb{R}^n$:
	\vspace{-0.1cm}
	\begin{align}\label{eq:monotone_flow_defn}
	\begin{aligned}
		F(x) = \left(\frac{\mathrm{Id}+G}{2}\right)^{-1}(x) - x, \\[-0.1cm]
	\end{aligned}
	\end{align}
	where $\mathrm{Id}$ denotes the identity function.
\end{defn}

Here, $F$ is well-defined since $(\mathrm{Id} + G)/2$ has a unique inverse by the Banach fixed point theorem~\cite{i-ResNets}. The definition of $F$ is a direct application of Theorem~\ref{thm:duality}. On the other hand, from the Cayley operator identity $C_{F}+C_{F^{-1}}=0$, we deduce that $F^{-1}$ has the Cayley operator $-G$. Hence, if $y=F(x)$,
\vspace{-0.1cm}
\begin{align}\label{eq:monotone_flow_inverse}
    \begin{aligned}
    x = \left(\frac{\mathrm{Id}-G}{2}\right)^{-1}(y) - y, \\[-0.1cm]
    \end{aligned}
\end{align}
which is well-defined by an analogous argument. Thus we arrive at the following theorem:

\begin{thm}
    For each $G$ with $\mathrm{Lip}(G) < 1$, there is a unique invertible function $F$ as in \eqref{eq:monotone_flow_defn}.
\end{thm}

Note that by rearranging \eqref{eq:monotone_flow_defn} or \eqref{eq:monotone_flow_inverse}, we obtain the implicit equation connecting $x$ and $y$:
\vspace{-0.1cm}
\begin{align}\label{eq:monotone_flow_implicit_eq}
    \begin{aligned}
	x-y = G(x+y). \\[-0.1cm]
	\end{aligned}
\end{align}
To parametrize $G$, we use the architecture proposed in i-DenseNets~\cite{i-DenseNets}, but replace the activation function with a new function called Concatenated Pila (CPila), which we introduce later in the paper.

\textbf{Remark.} While we parametrize each layer as a monotone operator, the resulting functions need not be monotone. 
For example, consider the $\pi/3$ counterclockwise rotation operator $R_{\pi/3}$ in $\mathbb{R}^2$. Although $R_{\pi/3}$ is monotone, the composition of $R_{\pi/3}$ with itself is not monotone. 
This illustrates that monotone functions are not closed under composition, hinting neural networks generated by the composition of invertible monotone operators can represent a much broader class of invertible functions than just monotone operators. Indeed, functions of the monotone formulation are $\sup$-universal diffeomorphism approximators since the monotone formulation includes the ``nearly-Id'' functions ($\mathcal{R}_L$ in section~\ref{sec:expressive_power}) whose composition suffices for $\sup$-universal diffeomorphism approximation~\cite{Teshima2020CFINN}.

\subsection{Training and inference}\label{sec:train_inf}

Training and inference largely follow previous approaches developed in  \cite{ResidualFlows,ImpFlows} except for minor modifications to adapt them to our monotone formulation. For the function $F$ of Definition~\ref{defn:monotone_flow}, implicit differentiation\footnote{We omit the absolute value sign since the strong monotonicity of $F$ implies $\det J_F>0$.} gives
\vspace{-0.1cm}
\begin{align}\label{eq:J_F}
    \begin{aligned}
	\log\det J_F = \mathrm{tr}\left\lbrack\log (I-J_G) - \log (I+J_G)\right\rbrack, \\[-0.1cm]
	\end{aligned}
\end{align}
where $J_G = \partial G(w)/\partial w$ at $w = \left(\frac{\mathrm{Id}+G}{2}\right)^{-1}(x)$; see Appendix~\ref{sec:derivation_J_F} for derivation. Now, we construct an unbiased estimator of the log determinant using the Hutchinson trace estimator and the unbiased Russian roulette estimator, following the approach of~\cite{ResidualFlows}. Combining (\ref{eq:change_of_variables}) and (\ref{eq:J_F}),
\begin{align}\label{eq:monotone_flow_logdet}
	\log p_X(x) = \log p_Z(z) + \mathbb{E}_{n\sim p_N(n), v\sim \mathcal{N}(0,I)}\left\lbrack\sum_{k=1}^{n}\frac{(-1) - (-1)^{k+1}}{k}\frac{v^T J_G^k v}{P(N\ge k)}\right\rbrack,
\end{align}
where $p_N(n)$ is a distribution with support over the positive integers. Note that the derivative of the log-likelihood (\ref{eq:monotone_flow_logdet}) can be evaluated by the memory-efficient Neumann gradient estimator as in~\cite{ResidualFlows}.

Training consists of forward and backward propagations using (\ref{eq:monotone_flow_defn}) and (\ref{eq:monotone_flow_logdet}). 
For forward propagation, $F$ can be evaluated with fixed-point iterations since $G$ is a contraction mapping while it suffices to backpropagate through $(\mathrm{Id}+G)^{-1}$ since $F(x) = \left(\frac{\mathrm{Id}+G}{2}\right)^{-1}(x) - x = (\mathrm{Id}+G)^{-1}(2x) - x$. For this, we follow the approach of~\cite{ImpFlows}. Let $G$ be parametrized by $\theta$. Denote the training objective (\ref{eq:monotone_flow_logdet}) as $\ell$ and write $w = (\mathrm{Id}+G)^{-1}(u)$ (where $u=2x$). Implicit differentiation yields
\begin{align}\label{eq:monotone_flow_implicit_diff_summary}
	\frac{\partial\ell}{\partial u} = \frac{\partial\ell}{\partial w}(I+J_G)^{-1}, \,\,\,\, \frac{\partial\ell}{\partial\theta} = \left(\frac{\partial\ell}{\partial w}(I+J_G)^{-1}\right) \frac{\partial G}{\partial\theta},
\end{align}
where $G$ and $J_G$ are evaluated at $w = (\mathrm{Id}+G)^{-1}(u)$; see Appendix~\ref{sec:derivation_implicit_diff} for derivation. Since the linear map $J_G: \mathbb{R}^n\rightarrow\mathbb{R}^n$ is a contraction mapping, the vector-matrix inverse product in \eqref{eq:monotone_flow_implicit_diff_summary} can be evaluated by a fixed-point iteration. For this, we adopt the secant method, but when convergence is not improved for 10 consecutive iterations, the Krasnoselskii-Mann iteration is used instead.

Inference consists of forward propagation through (\ref{eq:monotone_flow_inverse}), which is almost the same as the forward propagation in training except that the sign of $G$ is the opposite.

\subsection{Concatenated positive identity 1-Lipschitz activation function}

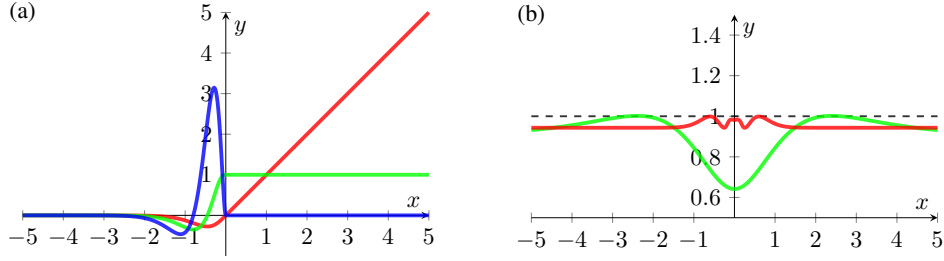
\begin{figure}
    \centering
    \vspace{-0.5cm}
    \hspace{-0.5cm}
    \begin{minipage}[b]{0.45\textwidth}
    	\centering
    	\begin{tikzpicture}[
    		declare function={
    			pila(\x)= ((\x)<=0) * (12.5*(\x)*(\x)*(\x)-5*(\x)*(\x)+(\x))*(exp(5*(\x))) + ((\x)>0) * (\x);
    			dpila(\x) = ((\x)<=0) * (1 - 5*(\x) + 12.5*(\x)*(\x) + 62.5*(\x)*(\x)*(\x)) * (exp(5*(\x))) + ((\x)>0) * (1);
    			ddpila(\x) = ((\x)<=0) * (312.5*(\x)*(\x)*(\x) + 250*(\x)*(\x))*(exp(5*(\x))) + ((\x)>0) * (0);
    		},
    		samples=1000,
    		scale=0.9
    		]
    		\begin{axis}[
    			axis x line=middle, axis y line=middle,
    			ymin=-1, ymax=5, ytick={0,...,5}, ylabel=$y$,y=0.6cm,
    			xmin=-5, xmax=5, xtick={-5,...,5}, xlabel=$x$,x=0.6cm,
    		    clip=false,
    			]
                \node (X) at (-5,5) {(a)};
    			\addplot[red, domain=-5:5, ultra thick, opacity=0.8]{pila(x)};
    			\addplot[green, domain=-5:5, ultra thick, opacity=0.8]{dpila(x)};
    			\addplot[blue, domain=-5:5, ultra thick, opacity=0.8]{ddpila(x)};
    		\end{axis}
    	\end{tikzpicture}
    \end{minipage}
    \hspace{0.3cm}
    \begin{minipage}[b]{0.45\textwidth}
    	\centering
    	\begin{tikzpicture}[
    		declare function={
    			dPila(\x) = ((\x)<=0) * (1 - 5*(\x) + 12.5*(\x)*(\x) + 62.5*(\x)*(\x)*(\x)) * (exp(5*(\x))) + ((\x)>0) * (1);
    			dLipSwish(\x) = (1/1.1) * ( 1 / (1+exp(-1*(\x))) + 1*(\x) * (1 / (1+exp(-1*(\x)))) * (1 / (1+exp(+1*(\x))));
    			speedCPila(\x) = sqrt(dPila((\x)-0.2)^2 + dPila(-(\x)-0.2)^2)/1.06;
    			speedCLipSwish(\x) = sqrt((dLipSwish(\x)^2 + dLipSwish(-(\x))^2)/1.004;
    		},
    		samples=1000,
    		scale=0.9
    		]
    		\begin{axis}[
    			axis x line=middle, axis y line=middle,
    			ymin=0.5, ymax=1.5, ylabel=$y$,y=3.0cm,
    			xmin=-5, xmax=5, xtick={-5,...,5}, xlabel=$x$,x=0.6cm,
    			clip=false
    			]
                \node (X) at (-5,1.5) {(b)};
    			\addplot[black, domain=-5:5, thick, dashed, opacity=0.8]{1};
    			\addplot[green, domain=-5:5, ultra thick, opacity=0.8]{speedCLipSwish(x)};
    			\addplot[red, domain=-5:5, ultra thick, opacity=0.8]{speedCPila(x)};
    		\end{axis}
    	\end{tikzpicture}
    \end{minipage}
    \caption{Graphical illustrations of Pila and CPila. (a) The graph of Pila (red) and its first (green) and second derivatives (blue) with $k=5$. (b) The speed of the curve of CPila (red) with $k=5$ and CLipSwish (green) with $\beta=1$.}
    \label{fig:pila}
\end{figure}

The LipSwish activation function proposed in~\cite{ResidualFlows} is based on the premise that having a non-vanishing second-order gradient would improve training. However, LipSwish still suffers from the vanishing gradient problem since its first-order derivative stays around 0.5 in the neighborhood of $x=0$ where most of the preactivations would reside. Concatenated LipSwish (CLipSwish)~\cite{i-DenseNets} improves upon LipSwish by concatenating LipSwish$(x)$ and LipSwish$(-x)$, but a similar issue still occurs as shown in Figure~\ref{fig:pila} (b).

In this regard, we propose Pila (positive identity 1-Lipschitz activation) function, which makes the gradient flow equal to the identity function on $x \ge 0$ during backpropagation.
When $x < 0$, Pila converges to zero as $x\to-\infty$ while matching the identity function at $x = 0$ up to the third derivative. 
The Pila function is described as follows:
\begin{align}\label{eq:Pila}
	\text{Pila}(x) =
	\begin{cases}
		x & \text{if $x \ge 0$}, \\
		\displaystyle
		\left(\frac{k^2}{2}x^3 - kx^2 + x\right)e^{kx} & \text{if $x < 0$}.
	\end{cases}
\end{align}

Here, $k > 0$ is treated either as a fixed hyperparameter or a learnable parameter. In all of our experiments, we set $k = 5$ and do not learn $k$. Note that $\text{Lip}(\text{Pila}) = 1$ for any $k > 0$.

As in \cite{i-DenseNets}, we introduce a concatenated version of Pila, which we call Concatenated Pila (CPila):
\vspace{-0.1cm}
\begin{align}\label{eq:CPila}
    \begin{aligned}
	\text{CPila}(x) = \alpha_1
		[\mathrm{Pila}(x-\alpha_2),
		\mathrm{Pila}(-x-\alpha_2)]^T, \\[-0.1cm]
	\end{aligned}
\end{align}
where $\alpha_1 = 1/1.06$ and $\alpha_2 = 0.2$. By rescaling and translating with $\alpha_1$ and $\alpha_2$, we fabricate the CPila function to be 1-Lipschitz. Figure~\ref{fig:pila} (b) describes the {\it speed} of the curve\footnote{speed of the curve at time $t$: $|v(t)| = \sqrt{(x'(t))^2+(y'(t))^2}$}, which shows CPila has a speed closer to $1$ around $x=0$ compared to the speed of CLipSwish.

\section{Expressive power}
    \label{sec:expressive_power}
\begin{figure}[t]
	\centering
	\vspace{-0.5cm}
	\begin{tikzpicture}[scale=0.4]
		\filldraw[color=red, fill=red!5, very thick, opacity=1.0](0,1.64/0.36) circle (1.64/0.36*1.6/1.64);
		\filldraw[color=green!40!gray, fill=green!5, very thick, opacity=0.75](0,1/0.36) circle (0.8/0.36);
		\filldraw[color=blue, fill=blue!5, very thick, opacity=0.5](0,1) circle (0.8);
		\draw[->] (-10,0)--(10,0) node[right]{$x$};
		\draw[->] (0,-0.2)--(0,9.5) node[above]{$y$};
        \draw[green!40!gray, dashed] (-5.5,1.8/0.36)--(+5.5,1.8/0.36) node[above]{$\frac{1}{1-L}$};
        \draw[thick, green!40!gray, dashed] (-5.5,0.2/0.36)--(+5.5,0.2/0.36) node[above]{$\frac{1}{1+L}$};
		\draw[blue, dashed] (-5.5,1.8)--(+8,1.8) node[above]{$\tiny{1+L}$};
		\draw[thick, blue, dashed] (-5.5,0.2)--(+8,0.2) node[above]{$\tiny{1-L}$};
		\draw[thick, red, dashed] (8,1.64/0.36*3.24/1.64)--(-8,1.64/0.36*3.24/1.64) node[below]{$\displaystyle\frac{1+L}{1-L}$};
		\draw[thick, red, dashed] (8,1.64/0.36*0.04/1.64)--(-8,1.64/0.36*0.04/1.64) node[above]{$\displaystyle\frac{1-L}{1+L}$};
		\node[red] at (1,8) {$\mathcal{M}_L$};
		\node[green!40!gray] at (0.82,4.1) {$\mathcal{I}_L$};
		\node[blue] at (0.2,1.1) {$\mathcal{R}_L$};
	\end{tikzpicture}
	\caption{Comparison of $\mathcal{R}_L$ (blue), $\mathcal{I}_L$ (green), and $\mathcal{M}_L$ (red). The figure depicts the possible ranges of $q = F(x_1) - F(x_2)$ with respect to $p = x_1 - x_2$ for each function class, assuming the domain and the codomain of $F$ are both $\mathbb{R}^2$. For visualization, we choose $L = 0.8$ and fix $p$ as a unit vector pointing in the $+y$ direction.}
    \label{fig:compare_RLILML}
\end{figure}
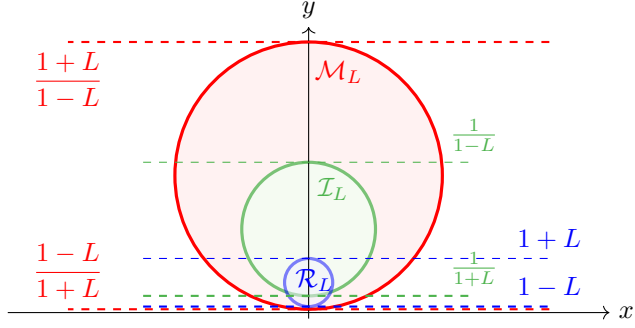

In this section, we analyze the expressive power of the monotone formulation and related models. We prove that the monotone formulation is superior to the residual formulation~\cite{ResidualFlows} and the implicit formulation of Lu et al.~\cite{ImpFlows} when the Lipschitz constant $L$ of the $G$-network is less than unity. In practice, since $L$ should not be close to unity, the monotone formulation is a better choice than the residual formulation or the implicit formulation of Lu et al.

We start with a formal definition of the function classes of each model.
\begin{defn}
	For $0\le L<1$, we define the following function classes:
	\vspace{-0.2cm}
	\begin{table}[H]
		\centering
		\setlength\tabcolsep{1.5pt}
		\begin{tabular}{lrl}
			$L$-Lipschitz functions & $\displaystyle \mathcal{G}_L$ & $\displaystyle = \left\lbrace G\in C^2(\mathbb{R}^n,\mathbb{R}^n) | \mathrm{Lip}(G) = L \right\rbrace$ \\
			Residual formulation & $\displaystyle \mathcal{R}_L$ & $\displaystyle = \left\lbrace \mathrm{Id}+G | G\in\mathcal{G}_L\right\rbrace$ \\
			Inverse residual formulation & $\displaystyle \mathcal{I}_L$ & $\displaystyle = \left\lbrace (\mathrm{Id}+G)^{-1}|G\in\mathcal{G}_L\right\rbrace$ \\
			Monotone formulation & $\displaystyle \mathcal{M}_L$ & $ = \left\lbrace \left(\frac{\mathrm{Id}+G}{2}\right)^{-1} - \mathrm{Id} \big| G\in\mathcal{G}_L\right\rbrace$
		\end{tabular}
	\end{table}
	\vspace{-0.6cm}
	\noindent Here, $C^2(\mathbb{R}^n,\mathbb{R}^n)$ denotes the set of twice continuously differentiable functions.\footnote{We require continuous second-order derivatives as the gradient of the log Jacobian determinant is of second-order. However, the results still hold when a weaker function class is considered (e.g., $C^1(\mathbb{R}^n,\mathbb{R}^n)$).}
\end{defn}
Note that the implicit formulation of Lu et al.~\cite{ImpFlows} defines one block as the composition of two functions $F_2^{-1}\circ F_1$ for any $F_1, F_2\in\mathcal{R}_L$. We perform our analysis only on the second layers, which corresponds to the comparison between $\mathcal{R}_L$ and $\mathcal{I}_L$. For convenience, we use the following notation.

\begin{defn}
	Let $A$ be one of $\mathcal{G}_L$, $\mathcal{R}_L$, $\mathcal{I}_L$, or $\mathcal{M}_L$, and $\alpha > 0$. Then,
	\begin{align*}
		\alpha A = \left\lbrace \alpha F | F\in A \right\rbrace, \,\,\,\,
		\mathbb{R}^+A = \left\lbrace \beta F | \forall\beta\in\mathbb{R}^+, F\in A\right\rbrace.
	\end{align*}
	\vspace{-0.7cm}
\end{defn}
We now examine the relations between the function spaces.
\begin{thm}\label{thm:relationship_between_function_classes}
	For $0\le L<1$, the following holds.
	\vspace{-0.2cm}
	\begin{center}
	(i) $\displaystyle \mathcal{I}_L = \frac{1}{1-L^2}\mathcal{R}_L$, \,\,\,\, (ii) $\displaystyle \mathcal{M}_L = \frac{1+L^2}{1-L^2}\mathcal{R}_{\frac{2L}{1+L^2}}$, \,\,\,\,
	(iii) $\mathcal{R}_L \subsetneq \mathcal{M}_L$, \,\,\,\,
	(iv) $\mathcal{I}_L \subsetneq \mathcal{M}_L$.
	\end{center}
\end{thm}
\textit{Proof.} See Appendix~\ref{apn:thm_relationship_between_function_classes} for a complete proof.

Figure~\ref{fig:compare_RLILML} shows the ranges of $q=F(x_1) - F(x_2)$ with respect to $p = x_1 - x_2$ for $\mathcal{R}_L$, $\mathcal{I}_L$, and $\mathcal{M}_L$, clearly demonstrating the relations (i)-(iv) of Theorem~\ref{thm:relationship_between_function_classes}.

Theorem~\ref{thm:relationship_between_function_classes} shows the inverse residual formulation $\mathcal{I}_L$ can only express functions that are a constant multiple of the functions expressed by the residual formulation $\mathcal{R}_L$. 
Hence, the inverse residual formulation has no advantage over the residual formulation in this respect. 
On the other hand, the monotone formulation $\mathcal{M}_L$ can express functions that are not expressible by any constant multiple of the functions in $\mathcal{R}_L$, because $2L/(1+L^2) > L$ whenever $L<1$.

When $L$ can approach unity arbitrarily close, the monotone formulation has the same expressive power as the residual formulation or the inverse residual formulation.
However, this is not possible in practice because the variance of the log determinant estimator (\ref{eq:monotone_flow_logdet}) diverges to infinity when $L$ is greater than a specific threshold less than unity that depends on the distribution $p_N(n)$~\cite{ContinuouslyIndexedFlow}. 
Since the log determinant estimator \eqref{eq:monotone_flow_logdet} is a linear combination of the log determinant estimator of the same $G$-network's residual formulation, \eqref{eq:monotone_flow_logdet} has a finite variance whenever the corresponding residual formulation's estimator possesses a finite variance. 
Suppose the constraints, in practice, bound the Lipschitz constant of the $G$-networks to be at most $L_\mathrm{max} < 1$. Then, the residual formulation gives rise to $\mathbb{R}^+\mathcal{R}_{L_\mathrm{max}}$ whereas the monotone formulation provides $\mathbb{R}^+\mathcal{M}_{L_\mathrm{max}} = \mathbb{R}^+\mathcal{R}_{\frac{2L_\mathrm{max}}{1+L_\mathrm{max}^2}} \supsetneq \mathbb{R}^+\mathcal{R}_{L_\mathrm{max}}$. 
An analogous argument holds for $\mathbb{R}^+\mathcal{I}_{L_\mathrm{max}}$. Hence, the analysis shows that the monotone formulation outperforms the other formulations in terms of the expressive power.

\section{Related work}
    \noindent\textbf{Normalizing flows.} Normalizing flows require each layer to be invertible and the Jacobian determinant of each layer to be tractable, which have spurred multiple approaches for architecture design. Restricted Jacobian models introduce specific structures to the Jacobian to make the Jacobian determinants tractable. 
They include determinant identity-based models~\cite{rezende2015variational, Woodbury} which use the Weinstein–Aronszajn identity with reduced intermediate layer dimensions, and coupling-based models~\cite{Nice, RealNVP, GLOW, FlowPlusPlus, Neuralsplineflows} which split the input into two parts and transform one part conditioned on another. However, the restrictions on the Jacobian limit their expressivity.
In contrast, free-form Jacobian models are not subject to the same constraints and include ResNet-based models (discussed in section~\ref{sec:prelim}) and Neural ODEs. 
Neural ODEs~\cite{FFJORD} parametrize the rate of change in ODEs with neural networks. They ensure invertibility through the existence and uniqueness theorem of ODEs. However, they have relatively high time complexity since they require solving ODEs numerically on every iteration.

So far, the normalizing flows we have discussed are non-autoregressive models. In contrast, autoregressive models~\cite{NAF,IAF,MAF} decompose probability distributions as conditional distributions depending only on the previous values. While autoregressive models typically achieve higher log-likelihoods, they are orders of magnitudes slower to train or sample~\cite{GLOW}. Therefore, we focus on non-autoregressive models in our paper.

\noindent\textbf{Monotone operators in neural networks.} \cite{monDEQ} pioneers monotone operator neural networks to assure the convergence of fixed-point iterations in \emph{implicit depth} networks. They build linear monotone operators by parametrizing the symmetric and antisymmetric parts separately. Positive defininteness is enforced only on the symmetric part. One shortcoming is that the construction is limited to \emph{linear} monotone operators, which our work addresses using nonlinear Cayley operators. \cite{Pesquet2021LMM} proposes the first neural network model of nonlinear MMOs by parametrizing the Cayley operators. One key difference is that the 1-Lipschitz constraint on Cayley operators is enforced through a penalty loss, which may fail to control the Lipschitz constant globally~\cite{i-ResNets}. Our method systematically ensures 1-Lipschitzness by the use of spectral normalization~\cite{SpectralNorm} and 1-Lipschitz activations, following~\cite{i-ResNets,ResidualFlows,i-DenseNets}. Also, 
their work does not consider normalizing flows.
To the best of our knowledge, our work proposes the first normalizing flow based on monotone operators.

\section{Experiments}
    We start by validating the modeling capacity of our model with a 1D toy function. We then evaluate Monotone Flows on density estimation of 2D toy and image datasets and conduct ablation studies. In all experiments, we parameterize the $G$-network in Definition~\ref{defn:monotone_flow} using i-DenseNets~\cite{i-DenseNets} with our activation function CPila (except for the 1D toy experiments that use CReLU and for part of the ablation study where we ablate CPila). The $G$-network consists of fully-connected layers for 1D and 2D toy tasks and convolutional layers for image tasks. We use the Poisson distribution for $p_N(n)$ whenever the estimator (\ref{eq:monotone_flow_logdet}) is used. We use Adam~\cite{Adam} for all experiments.

    \subsection{Toy 1D experiments}

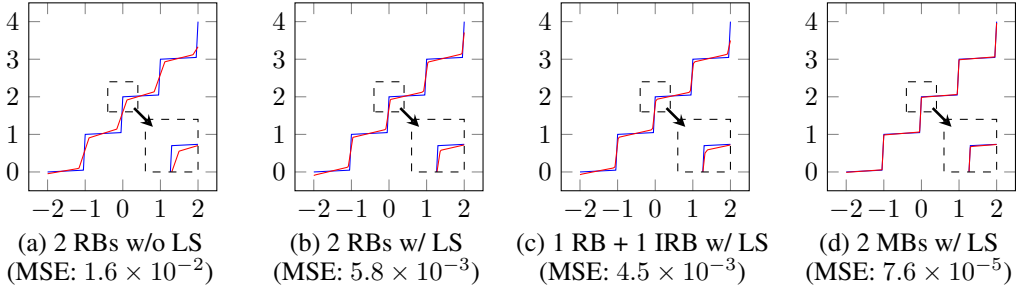
\begin{figure}
    \centering
    \vspace{-0.5cm}
    \captionsetup[sub]{font=normalsize}
    \begin{subfigure}[t]{0.24\textwidth}
    	\centering
    	\begin{tikzpicture}
    		\begin{axis}[
    			ymin=-0.5, ymax=4.5, ytick={0,...,4}, y=0.5cm,
    			xmin=-2.5, xmax=2.5, xtick={-2,...,2}, x=0.5cm,
    		    no markers,
                every axis plot/.append style={thin}
    			]
    			\addplot+ [blue] table [x=x_sample, y=y_sample, col sep=comma, sharp plot] {Figures/toy_1d/toy_1d_resflow_withoutActNorm_abridged.csv};
    			\addplot+ [red] table [x=x_sample, y=y_pred, col sep=comma, sharp plot] {Figures/toy_1d/toy_1d_resflow_withoutActNorm_abridged.csv};
    			\addplot+ [blue] table [x=x_sample_mag, y=y_sample_mag, col sep=comma, sharp plot] {Figures/toy_1d/toy_1d_magnified_gt.csv};
    			\addplot+ [red] table [x=x_sample_mag, y=y_pred_mag, col sep=comma, sharp plot] {Figures/toy_1d/toy_1d_resflow_withoutActNorm_magnified.csv};
    			\draw[dashed] (-0.4,1.6) -- (-0.4, 2.4) -- (0.4, 2.4) -- (0.4, 1.6) -- (-0.4, 1.6);
    			\draw[line width=1pt,black,-stealth](0.3,1.7)--(0.8,1.2) node[anchor=north west] {};
    			\draw[dashed] (0.6,1.4) -- (2.0,1.4) -- (2.0,0.0) -- (0.6,0.0) -- (0.6,1.4);
    		\end{axis}
    	\end{tikzpicture}
    	\vspace{-0.175cm}
    	\caption{2 RBs w/o LS \\ (MSE: $1.6\times10^{-2}$)}
    \end{subfigure}
    \hfill
    \begin{subfigure}[t]{0.24\textwidth}
    	\centering
    	\begin{tikzpicture}
    		\begin{axis}[
    			ymin=-0.5, ymax=4.5, ytick={0,...,4}, y=0.5cm,
    			xmin=-2.5, xmax=2.5, xtick={-2,...,2}, x=0.5cm,
    		    no markers,
                every axis plot/.append style={thin}
    			]
    			\addplot+ [blue] table [x=x_sample, y=y_sample, col sep=comma, sharp plot] {Figures/toy_1d/toy_1d_resflow_abridged.csv};
    			\addplot+ [red] table [x=x_sample, y=y_pred, col sep=comma, sharp plot] {Figures/toy_1d/toy_1d_resflow_abridged.csv};
    			\addplot+ [blue] table [x=x_sample_mag, y=y_sample_mag, col sep=comma, sharp plot] {Figures/toy_1d/toy_1d_magnified_gt.csv};
    			\addplot+ [red] table [x=x_sample_mag, y=y_pred_mag, col sep=comma, sharp plot] {Figures/toy_1d/toy_1d_resflow_magnified.csv};
    			\draw[dashed] (-0.4,1.6) -- (-0.4, 2.4) -- (0.4, 2.4) -- (0.4, 1.6) -- (-0.4, 1.6);
    			\draw[line width=1pt,black,-stealth](0.3,1.7)--(0.8,1.2) node[anchor=north west] {};
    			\draw[dashed] (0.6,1.4) -- (2.0,1.4) -- (2.0,0.0) -- (0.6,0.0) -- (0.6,1.4);
    		\end{axis}
    	\end{tikzpicture}
    	\vspace{-0.175cm}
    	\caption{2 RBs w/ LS \\ (MSE: $5.8\times10^{-3}$)}
    \end{subfigure}
    \hfill
    \begin{subfigure}[t]{0.24\textwidth}
    	\centering
    	\begin{tikzpicture}
    		\begin{axis}[
    			ymin=-0.5, ymax=4.5, ytick={0,...,4}, y=0.5cm,
    			xmin=-2.5, xmax=2.5, xtick={-2,...,2}, x=0.5cm,
    		    no markers,
                every axis plot/.append style={thin}
    			]
    			\addplot+ [blue] table [x=x_sample, y=y_sample, col sep=comma, sharp plot] {Figures/toy_1d/toy_1d_impflow_abridged.csv};
    			\addplot+ [red] table [x=x_sample, y=y_pred, col sep=comma, sharp plot] {Figures/toy_1d/toy_1d_impflow_abridged.csv};
    			\addplot+ [blue] table [x=x_sample_mag, y=y_sample_mag, col sep=comma, sharp plot] {Figures/toy_1d/toy_1d_magnified_gt.csv};
    			\addplot+ [red] table [x=x_sample_mag, y=y_pred_mag, col sep=comma, sharp plot] {Figures/toy_1d/toy_1d_impflow_magnified.csv};
    			\draw[dashed] (-0.4,1.6) -- (-0.4, 2.4) -- (0.4, 2.4) -- (0.4, 1.6) -- (-0.4, 1.6);
    			\draw[line width=1pt,black,-stealth](0.3,1.7)--(0.8,1.2) node[anchor=north west] {};
    			\draw[dashed] (0.6,1.4) -- (2.0,1.4) -- (2.0,0.0) -- (0.6,0.0) -- (0.6,1.4);
    		\end{axis}
    	\end{tikzpicture}
    	\vspace{-0.175cm}
    	\caption{1 RB + 1 IRB w/ LS \\ (MSE: $4.5\times10^{-3}$)}
    \end{subfigure}
    \hfill
    \begin{subfigure}[t]{0.24\textwidth}
        \centering
    	\begin{tikzpicture}
    		\begin{axis}[
    			ymin=-0.5, ymax=4.5, ytick={0,...,4}, y=0.5cm,
    			xmin=-2.5, xmax=2.5, xtick={-2,...,2}, x=0.5cm,
    		    no markers,
                every axis plot/.append style={thin}
    			]
    			\addplot+ [blue] table [x=x_sample, y=y_sample, col sep=comma, sharp plot] {Figures/toy_1d/toy_1d_mf_abridged.csv};
    			\addplot+ [red] table [x=x_sample, y=y_pred, col sep=comma, sharp plot] {Figures/toy_1d/toy_1d_mf_abridged.csv};
    			\addplot+ [blue] table [x=x_sample_mag, y=y_sample_mag, col sep=comma, sharp plot] {Figures/toy_1d/toy_1d_magnified_gt.csv};
    			\addplot+ [red] table [x=x_sample_mag, y=y_pred_mag, col sep=comma, sharp plot] {Figures/toy_1d/toy_1d_mf_magnified.csv};
    			\draw[dashed] (-0.4,1.6) -- (-0.4, 2.4) -- (0.4, 2.4) -- (0.4, 1.6) -- (-0.4, 1.6);
    			\draw[line width=1pt,black,-stealth](0.3,1.7)--(0.8,1.2) node[anchor=north west] {};
    			\draw[dashed] (0.6,1.4) -- (2.0,1.4) -- (2.0,0.0) -- (0.6,0.0) -- (0.6,1.4);
    		\end{axis}
    	\end{tikzpicture}
    	\vspace{-0.175cm}
    	\caption{2 MBs w/ LS \\ (MSE: $7.6\times10^{-5}$)}
    \end{subfigure}

	\caption{Comparison of 2 RBs without learnable scaling, 2 RBs, 1 RB followed by 1 IRB, and 2 MBs. 
	All experiments except (a) are performed with learnable scaling (LS). Blue and red lines represent the target function and the approximation by neural networks, respectively.} 
    \label{fig:toy_1d}
    
    \vspace{-0.3cm}
\end{figure}

Figure~\ref{fig:toy_1d} illustrates the expressive power of the monotone formulation with a 1D toy example.
We fit a 1D function that consists of four consecutive step-like shapes designed to exhibit a high Lipschitz constant. We consider four different 2-layer networks, as shown in Figure~\ref{fig:toy_1d}. Experimental details are discussed in Appendix~\ref{apn:experiment_toy_1d}. 
We denote residual block, inverse residual block, and monotone block by RB, IRB, and MB. Note that the learnable scaling (LS) is used only for Figure~\ref{fig:toy_1d} (b)-(d).
The $G$-networks in the experiments (a)-(d) have the same structure, and hence the same number of parameters are used. 
Consequently, differences in the result solely come from the differences in the formulations. 
Figure~\ref{fig:toy_1d} (a) and (b) demonstrate that the expressive power is improved when using $\mathbb{R}^+\mathcal{R}_L$ instead of $\mathcal{R}_L$. 
Figure~\ref{fig:toy_1d} (b) and (c) reveal that the expressive power of $\mathbb{R}^+\mathcal{R}_L$ and $\mathbb{R}^+\mathcal{I}_L$ are similar, in agreement with Theorem~\ref{thm:relationship_between_function_classes}.
Figure~\ref{fig:toy_1d} (d) shows that our monotone formulation $\mathbb{R}^+\mathcal{M}_L$ exceedingly outperforms the other networks.

    \subsection{Density estimation on 2D toy data}

\begin{figure}[t]
    \centering
    \hspace{-0.8cm}
	\begin{minipage}[b]{0.47\textwidth}
		\centering
        \captionsetup{width=.85\linewidth}
        \includegraphics[width=.85\linewidth]{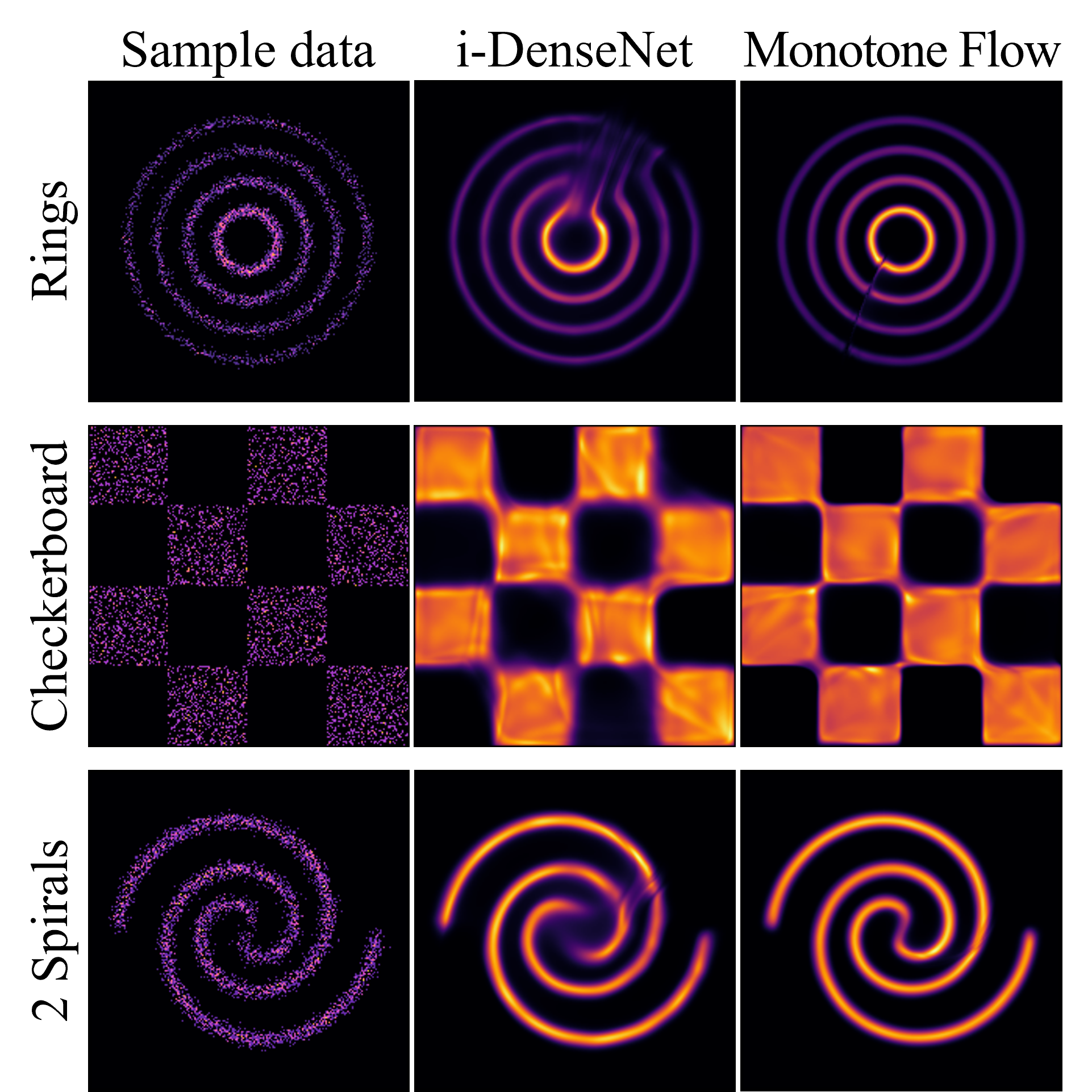}
        \captionof{figure}{2D toy density modeling results (full results in Appendix~\ref{apn:toy_full_results}).}
		\label{fig:fig_toy2d}
	\end{minipage}
    \begin{minipage}[b]{0.47\textwidth}
		\centering
		\captionof{table}{Toy density modelling results in nats. We display the average of the test loss for the last 20 tests at checkpoints (iterations 48100, 48200, ..., 50000) for a single run.}
	    \label{tab:tab_toy2d}
	    \setlength\tabcolsep{4 pt}
		\begin{tabular}{l|c|c}
            \hline
            Data & i-DenseNet & Monotone Flow\\
            \hline
            2 Spirals & 2.729 & \textbf{2.658} \\
            8 Gaussians & \textbf{2.840} & \textbf{2.840} \\
            Checkerboard & 3.609 & \textbf{3.540} \\
            Circles & 3.280 & \textbf{3.276} \\
            Moons & 2.401 & \textbf{2.400} \\
            Pinwheel & 2.343 & \textbf{2.333} \\
            Rings & 2.884 & \textbf{2.665} \\
            Swissroll & 2.680 & \textbf{2.676} \\
            \hline
        \end{tabular}
	    \setlength\tabcolsep{6 pt}
		\vspace{1.0cm}
	\end{minipage}
	\vspace{-0.5cm}
\end{figure}

We use the 2D toy datasets provided with the official source code of i-DenseNets~\cite{i-DenseNets}. Following~\cite{i-DenseNets}, we use ten flow blocks (see Appendix~\ref{apn:experiment_toy_2d} for details). Exact Jacobian evaluation is used instead of the stochastic estimation \eqref{eq:monotone_flow_logdet} as it is inexpensive for 2D data. We train the models for 50K epochs with the learnable concatenation of i-DenseNets enabled after 25K epochs. We use a batch size of 500.

Qualitative and quantitative results in Figure~\ref{fig:fig_toy2d} and Table~\ref{tab:tab_toy2d} show that Monotone Flows outperform i-DenseNets in almost all settings, especially on challenging datasets. One limitation of normalizing flows manifests in the qualitative results. For instance, in the dataset `rings', the circles cannot be closed because normalizing flows are required to preserve topology. Despite the inherent restriction, our model finds an excellent approximation.

    \subsection{Density estimation on images}
\label{sec:experiment_images}

\captionsetup[table]{skip=7pt}

\begin{table}[t]
    \centering
    \vspace{-12pt}
    \caption{Density estimation results on images in bits-per-dimension (bpd) with the number of parameters of each model. All numbers except for the last row are with uniform dequantization. VDQ: variational dequantization. }
    \label{tab:main}
    \begin{adjustbox}{width=\linewidth}
    \begin{tabular}{l cc cc cc cc}
        \toprule\midrule
        &
        \multicolumn{2}{c}{MNIST} & 
        \multicolumn{2}{c}{CIFAR-10} & 
        \multicolumn{2}{c}{ImageNet32} & 
        \multicolumn{2}{c}{ImageNet64}\\
        \cmidrule{2-9}
        Model & bpd $\downarrow$ & params & bpd $\downarrow$ & params & bpd $\downarrow$ & params & bpd $\downarrow$ & params \\ 
        \midrule
        Real NVP \cite{RealNVP} & 1.06 &  -    & 3.49 &  6.4M    & 4.28 & 46.0M & 3.98 & 96.0M\\
        Glow \cite{GLOW}         & 1.05 &  -    & 3.35 & 44.2M & 4.09 & 66.1M & 3.81 & 111.1M\\
        FFJORD \cite{FFJORD}  & 0.99 & - & 3.40 & - & - & - & - & -\\
        i-ResNet \cite{i-ResNets}      & 1.06 &  -    & 3.45 & 44.2M &  -   &   -   &  -   & -\\
        Residual Flow \cite{ResidualFlows} & 0.97 & 16.6M & 3.28 & 25.2M & 4.01 & 47.1M & 3.76 & 53.3M\\
        i-DenseNet \cite{i-DenseNets}    & -    &  -    & 3.25 & 24.9M & 3.98 & 47.0M &  -   & -\\
        \midrule
        Monotone Flow&\textbf{0.928}&20.9M&\textbf{3.215}&24.9M&\textbf{3.961}&47.0M&\textbf{3.734}&48.9M \\
        Monotone Flow + VDQ&-&-&\textbf{3.062}&46.9M&\textbf{3.901}&69.0M&-&- \\
        \bottomrule
    \end{tabular}
    \end{adjustbox}
\end{table}

We evaluate our method on MNIST~\cite{mnist}, CIFAR-10~\cite{cifar-10}, and the downscaled version of ImageNet in 32$\times$32 and 64$\times$64~\cite{PixelRNN}.\footnote{There are two versions of the downscaled ImageNet. The one used for evaluating normalizing flow models is the version packed in \emph{tar}. It has been removed from the official website but is still available through other routes (e.g., Academic Torrents).}
We use (\ref{eq:monotone_flow_logdet}) for estimating the Jacobian determinant. We use uniform dequantization for both training and testing. For MNIST and CIFAR-10, the learnable concatenation is enabled after 25 epochs; for ImageNet32 and ImageNet64, it is enabled from the start. We train for 100, 1,000, 20, 20 epochs with batch sizes 64, 64, 256, 256 with learning rates 0.001, 0.001, 0.004, 0.004 for MNIST, CIFAR-10, ImageNet32, ImageNet64, respectively. We report single-seed results following~\cite{GLOW, i-ResNets, ResidualFlows, i-DenseNets}. Details are in Appendix~\ref{apn:experiment_images}. Samples from the trained models are displayed in Figure~\ref{fig:fig_samples}. More samples can be found in Appendix~\ref{apn:image_samples}.

Table~\ref{tab:main} shows our model outperforms baselines on all datasets considered, demonstrating the effectiveness of our proposal. In concrete numbers, our model reduces 0.042 bpd in MNIST, 0.035 bpd in CIFAR-10, 0.019 bpd in ImageNet32, and 0.023 bpd in ImageNet64 compared to baseline normalizing flow models. 

\textbf{Remark on comparison.} Flow++ \cite{FlowPlusPlus} introduced variational dequantization which improves the log-likelihood, reporting 3.08 and 3.86 bits/dim on CIFAR-10 and ImageNet32, respectively. With variational dequantization, Monotone Flows achieve 3.062 and 3.901 bits/dim on the same benchmarks. Also, there is a line of research that achieves better likelihood values by introducing latent variables in extra dimensions, at the cost of losing exact likelihood computation \cite{VFlow, AugNF, DenselyConnectedNF}. On the other hand, ScoreFlow \cite{ScoreFlow} achieves better likelihood values by training with a weighted score-matching objective that lower bounds the log likelihood. We leave it for future research to adapt these methods for Monotone Flows.

\begin{figure*}[t]
    \centering
    \vspace{-0.1cm}
    \begin{subfigure}[t]{0.49\textwidth}
        \centering
        \includegraphics[width=\textwidth]{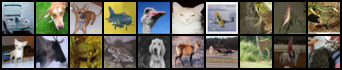}
        \vspace{-0.5cm}
        \caption{CIFAR-10 train data.}
        \vspace{0.1cm}
    \end{subfigure}
    \hfill
    \begin{subfigure}[t]{0.49\textwidth}
        \centering
        \includegraphics[width=\textwidth]{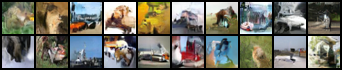}
        \vspace{-0.5cm}
        \caption{Monotone Flows trained on CIFAR-10.}
        \vspace{0.1cm}
    \end{subfigure}
    \begin{subfigure}[t]{0.49\textwidth}
        \centering
        \includegraphics[width=\textwidth]{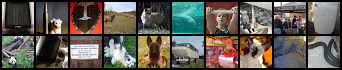}
        \vspace{-0.5cm}
        \caption{ImageNet32 train data.}
    \end{subfigure}
    \hfill
    \begin{subfigure}[t]{0.49\textwidth}
        \centering
        \includegraphics[width=\textwidth]{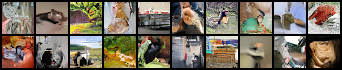}
        \vspace{-0.5cm}
        \caption{Monotone Flows trained on ImageNet32.}
    \end{subfigure}
    \caption{Train data and generated samples of CIFAR-10 and ImageNet32.}
    \label{fig:fig_samples}
    \vspace{-0.3cm}
\end{figure*}

    \subsection{Ablation study}

\captionsetup{width=.65\textwidth}

\begin{wraptable}[9]{r}{0.32\textwidth}
    \centering
    \vspace{-12pt}
    \caption{Ablation study for density estimation on CIFAR-10. MFL: monotone formulation.}
    \label{tab:ablation}
    \begin{tabular*}{0.32\textwidth}{@{\extracolsep{\fill}}cccccc}
         \hline
         & \# & MFL & CPila & bpd $\downarrow$ & \\
         \hline
         & 1 & \faTimes & \faTimes & 3.252 &\\
         & 2 & \faTimes & \faCheck & 3.243 &\\
         & 3 & \faCheck & \faTimes & 3.229 &\\
         & 4 & \faCheck & \faCheck & 3.215 &\\
         \hline
    \end{tabular*}
\end{wraptable}

Our model has two new components: the monotone formulation and CPila activation function. We quantify their contributions by ablating either of the two components while leaving the other intact. When ablating the monotone formulation (row \#2), we used the residual connections of \cite{i-DenseNets}; when ablating CPila (row \#3), we used CLipSwish instead. Table~\ref{tab:ablation} compares the performance of i-DenseNets (row \#1) and the ablated models on the CIFAR-10 dataset under the same experimental setup as in section~\ref{sec:experiment_images}. Details are in Appendix~\ref{apn:experiment_abl}. Results confirm that both components are beneficial for achieving a low bits-per-dimension (bpd).
The monotone formulation provides a higher performance gain (0.028 bpd) than CPila (0.014 bpd).

\section{Conclusion}
    \label{sec:conclusion}

We presented Monotone Flows, a novel parametrization of normalizing flows based on monotone operators (the monotone formulation) combined with a new activation function called Concatenated Pila (CPila). Our theoretical analysis elucidated why the monotone formulation is more expressive than the residual or inverse residual formulations. Experimental results demonstrated the effectiveness of the proposed method on various density estimation benchmarks. We believe our contribution is a solid step towards improving the expressivity of normalizing flows.

\begin{ack}
Our work is supported by ICT Creative Consilience program (IITP-2022-2020-0-01819) supervised by the IITP, and Kakao Brain Corporation. 
The work of Y. Hong was supported by Basic Science Research Program through the National Research Foundation of Korea (NRF) funded by the Ministry of Education (NRF-2021R1A2C1093579) and the Korea government (MSIT) (No. 2022R1A4A3033571).
\end{ack}

\bibliographystyle{unsrt}
\bibliography{reference}

\begin{enumerate}

\item For all authors...
\begin{enumerate}
    \item Do the main claims made in the abstract and introduction accurately reflect the paper's contributions and scope?
        \answerYes{}
    \item Did you describe the limitations of your work?
        \answerYes{In Appendix.}
    \item Did you discuss any potential negative societal impacts of your work?
        \answerYes{In Appendix.}
    \item Have you read the ethics review guidelines and ensured that your paper conforms to them?
        \answerYes{}
\end{enumerate}

\item If you are including theoretical results...
\begin{enumerate}
    \item Did you state the full set of assumptions of all theoretical results?
        \answerYes{}
    \item Did you include complete proofs of all theoretical results?
        \answerYes{In Appendix.}
\end{enumerate}

\item If you ran experiments...
\begin{enumerate}
    \item Did you include the code, data, and instructions needed to reproduce the main experimental results (either in the supplemental material or as a URL)?
        \answerYes{In the supplementary material.}
    \item Did you specify all the training details (e.g., data splits, hyperparameters, how they were chosen)?
        \answerYes{In Appendix.}
    \item Did you report error bars (e.g., with respect to the random seed after running experiments multiple times)?
        \answerNo{We use single-seed results following~\cite{GLOW,i-ResNets,ResidualFlows,i-DenseNets}.}
    \item Did you include the total amount of compute and the type of resources used (e.g., type of GPUs, internal cluster, or cloud provider)?
        \answerYes{In Appendix.}
\end{enumerate}

\item If you are using existing assets (e.g., code, data, models) or curating/releasing new assets...
\begin{enumerate}
    \item If your work uses existing assets, did you cite the creators?
        \answerYes{In section~\ref{sec:experiment_images}.}
    \item Did you mention the license of the assets?
        \answerNo{}
    \item Did you include any new assets either in the supplementary material or as a URL?
        \answerNA{}
    \item Did you discuss whether and how consent was obtained from people whose data you're using/curating?
        \answerNA{}
    \item Did you discuss whether the data you are using/curating contains personally identifiable information or offensive content?
        \answerNo{}
\end{enumerate}

\item If you used crowdsourcing or conducted research with human subjects...
\begin{enumerate}
    \item Did you include the full text of instructions given to participants and screenshots, if applicable?
        \answerNA{}
    \item Did you describe any potential participant risks, with links to Institutional Review Board (IRB) approvals, if applicable?
        \answerNA{}
    \item Did you include the estimated hourly wage paid to participants and the total amount spent on participant compensation?
        \answerNA{}
\end{enumerate}

\end{enumerate}

\clearpage


\appendix

{
\hrule height 4pt
  \vskip 0.25in
  \vskip -\parskip

\centering
    {\LARGE\bf Invertible~Monotone~Operators~for~Normalizing~Flows (Supplementary Material) \par}

\vskip 0.29in
  \vskip -\parskip
  \hrule height 1pt
  \vskip 0.09in
 
\vskip 0.5cm
}

\textbf{Overview.} The Appendix is organized as follows. In Appendix A, we present the proofs of the theorems and lemmas stated in the main text. 
In Appendix B, we present the forward and backward algorithms for a single monotone layer. 
In Appendix C, we describe the experimental details for each experiment discussed in the main text and present the classification experiment on CIFAR-10. In Appendix D and E, we include the visualizations of the 2D toy experiments and the generated images from our trained model. In Appendix F, we discuss the limitations and potential negative societal impacts of our work.

\section{Proofs and derivations}
    \subsection{Proof of the strong monotonicity of a single residual block}\label{apn:residual_formulation_strongly_monotone}

We prove that $R(x) = x + G(x)$ with $\mathrm{Lip}(G) = L$ is $(1-L)$-strongly monotone. 
A direct calculation shows
\begin{align*}
	\langle R(x)-R(y),x-y\rangle &= \lVert x-y\rVert_2^2 - \langle G(x)-G(y),x-y\rangle \\
	&\ge \lVert x-y\rVert_2^2 - \lVert G(x)-G(y)\rVert_2 \lVert x-y\rVert_2 \\
	&\ge \lVert x-y\rVert_2^2 - L\lVert x-y\rVert_2 \lVert x-y\rVert_2 = (1-L)\lVert x-y\rVert_2^2.
\end{align*}

\subsection{Proof of Theorem~\ref{thm:duality}}

As stated in the main text, a complete proof can be found in~\cite[Proposition 23.8 and Proposition 4.4]{bauschke2011convex}. 
Here, we provide an alternative proof, largely based on the proof of Alberti and Ambrosio \cite{Alberti1999AGA}, to keep our paper self-contained.
Before proving Theorem~\ref{thm:duality}, we state Kirszbraun's theorem \cite{Kirszbraun1934}.

\begin{thm}\label{thm:Kirszbraun}
	(Kirszbraun, 1934) If $U\subseteq\mathbb{R}^n$ and $f: U\rightarrow\mathbb{R}^m$ is $K$-Lipschitz, there is a $K$-Lipschitz function $g: \mathbb{R}^n\rightarrow\mathbb{R}^m$ which is an extension of of $f$ to $\mathbb{R}^n$.
\end{thm}

Now we prove Theorem~\ref{thm:duality}.

\textit{Proof.} (i) $(\Rightarrow)$ Assume $F$ is monotone, and let $(x_1,y_1),(x_2,y_2)\in C_F$.
By rearranging the equation of $C_F$, we obtain
\begin{align*}
    b_i = F(a_i),\quad 
    a_i = \frac{x_i+y_i}{2}, \quad 
    b_i = \frac{x_i-y_i}{2}, \quad
    \text{for  } i=1, 2.
\end{align*}
Hence, we find that 
\begin{align*}
    \left\lVert y_1-y_2\right\rVert_2^2 &= \left\lVert \left(a_1-b_1\right) - \left( a_2-b_2\right) \right\rVert_2^2 \\
    &= \left\lVert \left(a_1-a_2\right) - \left( b_1-b_2\right) \right\rVert_2^2 \\
    &= \left\lVert a_1 - a_2 \right\rVert_2^2 + \left\lVert b_1 - b_2 \right\rVert_2^2 - 2\left\langle a_1 - a_2, b_1 - b_2\right\rangle \\
    &\le \left\lVert a_1 - a_2 \right\rVert_2^2 + \left\lVert b_1 - b_2 \right\rVert_2^2 + 2\left\langle a_1 - a_2, b_1 - b_2\right\rangle \\
    &= \left\lVert \left(a_1-a_2\right) + \left(b_1-b_2\right)\right\rVert_2^2 \\
    &= \left\lVert \left(a_1+b_1\right) - \left(a_2+b_2\right)\right\rVert_2^2 \\
    &= \left\lVert x_1 - x_2 \right\rVert_2^2,
\end{align*}
where the inequality follows from the monotonicity of $F$. Hence, $C_F$ is 1-Lipschitz.

$(\Leftarrow)$ Assume $C_F$ is 1-Lipschitz, and let $(x_1,y_1),(x_2,y_2)\in F$. 
Rearranging the equation of $C_F$, we deduce that 
\begin{align*}
    d_i = C_F(c_i), \quad
    c_i = x_i+y_i, \quad
    d_i = x_i-y_i, \quad
    \text{for  } i=1, 2.
\end{align*}
Hence, we derive that
\begin{align*}
    \langle y_1-y_2,x_1-x_2\rangle &= \left\langle \frac{c_1-d_1}{2}-\frac{c_2-d_2}{2},\frac{c_1+d_1}{2}-\frac{c_2+d_2}{2}\right\rangle \\
    &= \left\langle \frac{c_1-c_2}{2}-\frac{d_1-d_2}{2},\frac{c_1-c_2}{2}+\frac{d_1-d_2}{2}\right\rangle \\
    &= \frac{1}{4}\bigg( \left\lVert c_1-c_2\right\rVert_2^2 - \left\lVert d_1-d_2\right\rVert_2^2 \bigg) \ge 0,
\end{align*}
where the last inequality follows from the 1-Lipschitz condition of $C_F$. Hence, $F$ is monotone.

(ii) $(\Rightarrow)$ Assume $F$ is maximally monotone. Then, $C_F$ is 1-Lipschitz by (i), whence $C_F$ is at most single-valued at each point. 
Suppose $\mathrm{dom}\,C_F\subsetneq\mathbb{R}^n$. 
By Kirszbraun's theorem (Theorem~\ref{thm:Kirszbraun}), there is an extension $F_e$ of $C_F$ such that $F_e$ is 1-Lipschitz and $\mathrm{dom}\, F_e = \mathbb{R}^n$. 
Since $F_e$ is 1-Lipschitz, the operator $\tilde{F}$ that has $F_e$ as its Cayley operator is monotone. We have $F\subsetneq \tilde{F}$, which contradicts the maximality of $F$. 

$(\Leftarrow)$ Assume $C_F$ is 1-Lipschitz and $\mathrm{dom}\,C_F=\mathbb{R}^n$. 
Then, $F$ is monotone by (i). Suppose $F$ is not maximally monotone. This implies there exists an $(x^\prime,y^\prime)\in\mathbb{R}^n\times\mathbb{R}^n$ such that $(x^\prime,y^\prime)\notin F$ but $\widehat{F} := F\cup\lbrace(x^\prime,y^\prime)\rbrace$ is monotone. Thus, the Cayley operator of $\widehat{F}$,
\begin{align*}
    C_{\widehat{F}} = C_F \cup \left\lbrace \left(x^\prime+y^\prime, x^\prime-y^\prime \right) \right\rbrace,
\end{align*}
is 1-Lipschitz by (i). 
On the other hand, $C_{\widehat{F}}$ has a function value $C_F(x^\prime+y^\prime)$ at $x^\prime+y^\prime$. 
This does not equal $x^\prime-y^\prime$ since $(x^\prime,y^\prime) \notin F$. 
This means $C_{\widehat{F}}$ is multi-valued at $x^\prime+y^\prime$, which contradicts $C_{\widehat{F}}$ is 1-Lipschitz. \qed

\subsection{Derivation of equation (\ref{eq:J_F})}\label{sec:derivation_J_F}

As mentioned in the main text, we assume $G$ is continuously twice differentiable and is a contraction mapping with $\mathrm{Lip}(G) = L < 1$. 
The Jacobian of $F$ can be calculated in a straightforward manner from the explicit form~\eqref{eq:monotone_flow_defn}:
\begin{align*}
    J_F = \frac{\partial y}{\partial x} = J_{\left(\frac{\mathrm{Id}+G}{2}\right)^{-1}} - I = 2J_{(\mathrm{Id}+G)^{-1}}-I.
\end{align*}
Here, we have $\lVert J_G\rVert_2 \le \mathrm{Lip}(G) = L$, where $\lVert\cdot\rVert$ denotes the matrix spectral norm. 
Thus, for any $v\in\mathbb{R}^n$ with $\lVert v\rVert_2 = 1$, we have $v^T J_{\mathrm{Id}+G} v = v^T v + v^T J_G v\ge 1-L>0$.
This implies $J_{\mathrm{Id}+G}$ is not singular and thus invertible. 
By inverse function theorem, we obtain that  $J_{(\mathrm{Id}+G)^{-1}} = J_{\mathrm{Id}+G}^{-1}$. Therefore, we deduce that
\begin{align*}
	J_F &= 2J_{(\mathrm{Id}+G)^{-1}} - I \\
	&= 2J_{\mathrm{Id}+G}^{-1} - I \\
	&= 2(I+J_G)^{-1} - I \\
	&= (I+J_G)^{-1}(2I-(I+J_G)) \\
	&= (I+J_G)^{-1}(I-J_G),
\end{align*}
where $J_G$ is evaluated at $w = (\mathrm{Id}+G)^{-1}(u)$ and $u=2x$. 
Hence, we find that
\begin{align*}
	\log\det J_F &= \log\det \left\lbrack (I+J_G)^{-1}(I-J_G) \right\rbrack \\
	&= \log\det(I-J_G) - \log\det(I+J_G) \\
	&= \mathrm{tr} \left\lbrack \log(I-J_G) - \log(I+J_G) \right\rbrack
\end{align*}
where the function $\log$ in the last line denotes the matrix logarithm (not an elementwise logarithm). 
In our implementation, we use $G(x) = -2H(x/\sqrt{2})$ where $H$ is an $L$-Lipschitz function. 
This change does not affect the Lipschitz constant in the calculation, and transforms the formulation of $F$ to $F(x) = \sqrt{2}(\mathrm{Id}-H)^{-1}(\sqrt{2}x)-x$. 
The inverse of $F$ can be derived by $F^{-1}(x) = \sqrt{2}(\mathrm{Id}+H)^{-1}(\sqrt{2}x)-x$ as in \eqref{eq:monotone_flow_inverse}.

\subsection{Derivation of equation (\ref{eq:monotone_flow_logdet})}

By Taylor expansion, we derive that
\begin{align*}
	\log\det J_F = \mathrm{tr} \left\lbrack \log(I-J_G) - \log(I+J_G) \right\rbrack = \mathrm{tr} \left\lbrack \sum_{k=1}^{\infty} \frac{(-1)-(-1)^{k+1}}{k} J_G^k \right\rbrack.
\end{align*}
Here, Taylor expansion is justified because $||J_G||_2 \le \mathrm{Lip}(G) = L < 1$. Now, with neural networks, the exact evaluation of $J_G^k$'s is not feasible due to computational complexity. 
Instead, we resort to an estimator called the Hutchinson trace estimator, which only requires the vector-matrix or matrix-vector product. 
Given a general matrix $A\in\mathbb{R}^{n\times n}$, the Hutchinson trace estimator is defined as
\begin{align*}
	\mathrm{tr}(A) = \mathbb{E}_{v\sim\mathcal{N}(0,I)}\left\lbrack v^T Av\right\rbrack,
\end{align*}
where $v$ is sampled from a multivariate standard normal distribution. 
When applied to our case, this yields
\begin{align*}
	\log\det J_F = \mathbb{E}_{v\sim \mathcal{N}(0, I)} \, \left\lbrack \sum_{k=1}^{\infty} \frac{(-1)-(-1)^{k+1}}{k} v^T J_G^k v \right\rbrack.
\end{align*}
The calculation contains an infinite series whose evaluation is difficult. 
However, the infinite sum can be approximated by a finite number of terms using the unbiased Russian roulette estimator:
\begin{align*}
	\log\det J_F = \mathbb{E}_{n\sim p_N(n),v\sim \mathcal{N}(0, I)} \,\left\lbrack \sum_{k=1}^{n} \frac{(-1)-(-1)^{k+1}}{k} \frac{v^T J_G^k v}{P(N\ge k)} \right\rbrack.
\end{align*}
Here, the distribution $p_N(n)$ can be any distribution with $P(N\ge k) > 0$ for all natural numbers $k\in\mathbb{N}$; for instance, a geometric distribution is chosen in \cite{ResidualFlows}, and a Poisson distribution is chosen in \cite{i-DenseNets}. 
In our case, we use the Poisson distribution following~\cite{i-DenseNets}.

The Russian roulette estimator is justified by Lemma 3 in Appendix B of~\cite{ResidualFlows}.
More precisely, we deduce that
\begin{align*}
	\sum_{k=1}^{\infty} \left\lvert \frac{(-1)-(-1)^{k+1}}{k} v^T J_G^k v\right\rvert
	&\le 2\sum_{k=1}^{\infty} \left\lvert \frac{1}{k} v^T J_G^k v \right\rvert \\
	&\le 2\sum_{k=1}^{\infty} \frac{1}{k} \norm{v}_2 \norm{J_G}_2^k \norm{v}_2 \\
	&= 2 \norm{v}_2^2 \sum_{k=1}^{\infty} \frac{1}{k}\norm{J_G}_2^k \\
	&\le 2 \norm{v}_2^2 \sum_{k=1}^{\infty} \frac{1}{k}\mathrm{Lip}(G)^k \\
	&= 2 \norm{v}_2^2 \log(1-\mathrm{Lip}(G)) < \infty.
\end{align*}

We include the following lemma here to keep our paper self-contained.
\begin{lemma}
    (Lemma 3 in~\cite{ResidualFlows}) (Unbiased randomized truncated series) Let $Y_k$ be a real random variable with $\lim_{k\to\infty}\mathbb{E}\lbrack Y_k\rbrack = a$ for some $a\in\mathbb{R}$. Further, let $\Delta_0=Y_0$ and $\Delta_k=Y_k-Y_{k-1}$ for $k\ge 1$. Assume $\mathbb{E}\left\lbrack \sum_{k=0}^{\infty}|\Delta_k|\right\rbrack<\infty$ and let $N$ be a random variable with support over the positive integers and $n\sim p_N(n)$. Then for
    \begin{align*}
        Z=\sum_{k=0}^{n} \frac{\Delta_k}{P(N\ge k)},
    \end{align*}
    we find
    \begin{align*}
        \lim_{k\to\infty}\mathbb{E}\left\lbrack Y_k\right\rbrack = \mathbb{E}_{n\sim p_N(n)}\left\lbrack Z\right\rbrack = a.
    \end{align*}
\end{lemma}

In our case, $\Delta_0 = 0$ and for $k\ge 1$
\begin{align*}
    \Delta_k = \frac{((-1)-(-1)^{k+1})}{k}v^T J_G^k v.
\end{align*}

\subsection{Derivation of equation (\ref{eq:monotone_flow_implicit_diff_summary})}\label{sec:derivation_implicit_diff}

Backward propagation is done through implicit differentiation by adapting the formulation in~\cite{ImpFlows}. For completeness, we include the derivation here.

Since $F(x) = \left(\frac{\mathrm{Id}+G}{2}\right)^{-1}(x) - x = (\mathrm{Id}+G)^{-1}(2x) - x$, it suffices to backpropagate through the function $(\mathrm{Id}+G)^{-1}$. 
Let $w = (\mathrm{Id}+G)^{-1}(u)$ (with $u=2x$), where $G$ is parameterized with parameters $\theta$. 
Here, $u$ and $\theta$ are independent variables, and $w$ and the loss $\ell$ are dependent variables. 
We first note that by chain rule
\begin{align*}
	\frac{\partial\ell}{\partial u} = \frac{\partial\ell}{\partial w}\frac{\partial w}{\partial u},\,\,\,\,\frac{\partial\ell}{\partial\theta} = \frac{\partial\ell}{\partial w}\frac{\partial w}{\partial\theta}.
\end{align*}
Since $\partial\ell/\partial y$ is given by backpropagation, we only need to estimate the vector-Jacobian products $(\partial\ell/\partial w)(\partial w/\partial u)$ and $(\partial\ell/\partial w)(\partial w/\partial\theta)$ from $\partial\ell/\partial y$, not the full Jacobians $\partial w/\partial u$ and $\partial w/\partial\theta$.

To find $\partial\ell/\partial u$, we consider the implicit equation between $u$ and $w$ given by
\begin{align*}
	w + G(w, \theta) - u = 0.
\end{align*}
Taking a derivative with respect to $w$ while holding $\theta$ constant yields
\begin{align*}
	\frac{\partial w}{\partial u} + J_G \frac{\partial w}{\partial u} - I = 0,
\end{align*}
where $J_G \equiv (\partial G(x, \theta)/\partial x)|_{x=w,\,\theta=\theta}$. Hence, we obtain
\begin{align}\label{eq:apn_fixed_point_backward}
	\frac{\partial\ell}{\partial u} = \frac{\partial\ell}{\partial w}(I+J_G)^{-1} \,\,\,\,\Rightarrow\,\,\,\,\frac{\partial\ell}{\partial u} (I+J_G) = \frac{\partial\ell}{\partial w}.
\end{align}
The equation on the right of \eqref{eq:apn_fixed_point_backward} can be solved for $\partial\ell/\partial u$ using fixed-point iterations since $\mathrm{Lip}(J_G) \le L < 1$.
We now consider $\partial\ell/\partial\theta$. 
Taking differentiation with respect to $\theta$ while keeping $u$ constant yields
\begin{align*}
	\frac{\partial w}{\partial\theta} + \left(J_G\frac{\partial w}{\partial\theta} + \frac{\partial G(x, \theta)}{\partial\theta}\bigg|_{x=w,\,\theta=\theta}\right) - 0 = 0.
\end{align*}
Hence, we deduce that
\begin{align} \label{app:11}
	\frac{\partial\ell}{\partial\theta} = (-1)\left(\frac{\partial\ell}{\partial w} (I+J_G)^{-1}\right) \frac{\partial G(x,\theta)}{\partial\theta}\bigg|_{x=w,\,\theta=\theta}.
\end{align}
Since the term in parentheses is the same as $\partial\ell/\partial u$, we can reuse the result from the fixed-point iteration for~\eqref{eq:apn_fixed_point_backward}. 
Notice that \eqref{app:11} is similar to the form of 
\begin{align*}
	\frac{\partial\ell}{\partial\theta} = \frac{\partial\ell}{\partial G(\mathrm{stop\_gradient}(w),\theta)}\frac{\partial G(\mathrm{stop\_gradient}(w),\theta)}{\partial\theta}.
\end{align*}
Hence, we can backpropagate through $G$ using the standard backpropagation approach by setting the output gradient as $\partial\ell/\partial u$.

\subsection{Proof of Theorem \ref{thm:relationship_between_function_classes}}\label{apn:thm_relationship_between_function_classes}

\textit{Proof.} We start by noting that the functions in $\mathcal{R}_L$ are continuously differentiable by construction. In fact, the functions in $\mathcal{I}_L$ are continuously differentiable by the inverse function theorem since the Jacobian $I+J_G$ of $\mathrm{Id}+G$ is nonsingular. 
Moreover, the functions in $\mathcal{M}_L$ are continuously differentiable since they can be written in terms of the functions of $\mathcal{I}_L$.
When the function $F$ such that $F: \mathbb{R}^n\rightarrow\mathbb{R}^n$ is invertible, the inverse function theorem implies that $F$ and $F^{-1}$ share the same differentiability, and so do $\mathcal{G}_L$, $\mathcal{R}_L$, $\mathcal{I}_L$, and $\mathcal{M}_L$.
Hence, it suffices to consider the Lipschitz condition. For $x_1,x_2\in\mathbb{R}^n$, write $p = x_1 - x_2$ and $q = F(x_1) - F(x_2)$. 
Then, we find that
\begin{align*}
	F\in\mathcal{R}_L &\Leftrightarrow F\in C^2(\mathbb{R}^n,\mathbb{R}^n), \forall x_1,x_2\in\mathbb{R}^n \,\, \lVert q-p\rVert_2 \le L\lVert p\rVert_2, \\
	F\in\mathcal{I}_L &\Leftrightarrow F\in C^2(\mathbb{R}^n,\mathbb{R}^n), \forall x_1,x_2\in\mathbb{R}^n \,\, \lVert p-q\rVert_2 \le L\lVert q\rVert_2 \\
	&\overset{(*)}{\Leftrightarrow} F\in C^2(\mathbb{R}^n,\mathbb{R}^n), \forall x_1,x_2\in\mathbb{R}^n \,\, \lVert(1-L^2)q-p\rVert_2 \le L\lVert p\rVert_2 \\
	&\Leftrightarrow (1-L^2)F\in \mathcal{R}_L \\
	&\Leftrightarrow F\in \frac{1}{1-L^2}\mathcal{R}_L, \\
	F\in\mathcal{M}_L &\Leftrightarrow F\in C^2(\mathbb{R}^n,\mathbb{R}^n), \forall x_1,x_2\in\mathbb{R}^n \,\, \lVert q-p\rVert_2 \le L\lVert q+p\rVert_2 \\
	&\overset{(**)}{\Leftrightarrow} F\in C^2(\mathbb{R}^n,\mathbb{R}^n), \forall x_1,x_2\in\mathbb{R}^n \,\, \left\lVert\left(\frac{1-L^2}{1+L^2}\right)q-p\right\rVert_2 \le \left(\frac{2L}{1+L^2}\right)\lVert p\rVert_2 \\
	&\Leftrightarrow \left(\frac{1-L^2}{1+L^2}\right)F\in\mathcal{R}_{\frac{2L}{1+L^2}} \\
	&\Leftrightarrow F\in \left(\frac{1+L^2}{1-L^2}\right)\mathcal{R}_{\frac{2L}{1+L^2}}.
\end{align*}
Note that the equivalence (*) can be derived as follows:
\begin{align*}
    &\lVert p-q\rVert_2 \le L\lVert q\rVert_2 \\
    &\Leftrightarrow \lVert p-q\rVert_2^2 \le L^2\lVert q\rVert_2^2 \\
    &\Leftrightarrow \lVert p\rVert_2^2 -2\langle p,q\rangle + (1-L^2)\lVert q\rVert_2^2 \le 0 \\
    &\Leftrightarrow (1-L^2)^2\lVert q\rVert_2^2 - 2(1-L^2)\langle p,q\rangle + (1-L^2)\lVert p\rVert_2^2 \le 0 \\
    &\Leftrightarrow \lVert (1-L^2)q-p\rVert_2^2 \le L^2\lVert p\rVert_2^2 \\
    &\Leftrightarrow \lVert (1-L^2)q-p\rVert_2 \le L\lVert p\rVert_2.
\end{align*}
Also, the equivalence (**) can be calculated as follows:
\begin{align*}
    &\lVert q-p\rVert_2 \le L\lVert p+q\rVert_2 \\
    &\Leftrightarrow \lVert q-p\rVert_2^2 \le L^2\lVert p+q\rVert_2^2 \\
    &\Leftrightarrow (1-L^2)\lVert p\rVert_2^2 - 2(1+L^2)\langle p,q\rangle + (1-L^2)\lVert q\rVert_2^2 \le 0 \\
    &\Leftrightarrow (1-L^2)^2\lVert q\rVert_2^2 - 2(1-L^2)(1+L^2)\langle p,q\rangle + (1-L^2)^2\lVert p\rVert_2^2 \le 0 \\
    &\Leftrightarrow \lVert (1-L^2)q - (1+L^2)p\rVert_2^2 \le 4L^2\lVert p\rVert_2^2 \\
    &\Leftrightarrow \left\lVert \frac{1-L^2}{1+L^2}q - p\right\rVert_2 \le \frac{2L}{1+L^2}\lVert p\rVert_2.
\end{align*}
Hence, the statements (i) and (ii) are proved.

Since $\mathcal{I}_L$ and $\mathcal{M}_L$ are explicitly characterized in terms of $\mathcal{R}_L$ (or $\mathcal{R}_\frac{2L}{1+L^2}$), we utilize this fact to show the statements (iii) and (iv).
For (iii), suppose $F\in\mathcal{R}_L$. 
Then, $F = \mathrm{Id} + G$ for some $G\in\mathcal{G}_L$. 
Since
\begin{align*}
    F = \mathrm{Id} + G = \frac{1+L^2}{1-L^2}\left\lbrack \mathrm{Id} + \left( \frac{-2L^2}{1+L^2}\mathrm{Id} + \frac{1-L^2}{1+L^2}G\right)\right\rbrack
\end{align*}
and
\begin{equation*}
\begin{split}
    \mathrm{Lip}\left(\frac{-2L^2}{1+L^2}\mathrm{Id} + \frac{1-L^2}{1+L^2}G\right) 
    & \le \frac{2L^2}{1+L^2} + \frac{1-L^2}{1+L^2}L \\
    & = \frac{2L}{1+L^2}\cdot\left(\frac{1+2L-L^2}{2}\right) \\
    & < \text{(since } 0\le L<1)\\
    & < 
    \frac{2L}{1+L^2},
\end{split}
\end{equation*}
we then have $F\in\mathcal{M}_L$.

For (iv), suppose $F\in\mathcal{I}_L$. Then, $\displaystyle F = \frac{1}{1-L^2}(\mathrm{Id} + G)$ for some $G\in\mathcal{G}_L$. Since
\begin{align}
    F = \frac{1}{1-L^2}(\mathrm{Id}+G) = \frac{1+L^2}{1-L^2}\left\lbrack \mathrm{Id} + \left(\frac{-L^2}{1+L^2}\mathrm{Id}+\frac{1}{1+L^2}G\right)\right\rbrack
\end{align}
and
\begin{align*}
    \mathrm{Lip}\left(\frac{-L^2}{1+L^2}\mathrm{Id}+\frac{1}{1+L^2}G\right) \le \frac{L^2}{1+L^2} + \frac{1}{1+L^2}{L} = \frac{2L}{1+L^2}\cdot\left(\frac{1+L}{2}\right)<\frac{2L}{1+L^2},
\end{align*}
we obtain that $F\in\mathcal{M}_L$.
We finally notice that the above derivation, indeed, shows that the Lipschitz constant of the residual part $\mathrm{Lip}\left(\frac{1-L^2}{1+L^2}F - \mathrm{Id}\right)$ is always \emph{smaller} than $\frac{2L}{1+L^2}$ for functions $F$ in $\mathcal{R}_L$ or $\mathcal{I}_L$. 
Hence, both $\mathcal{R}_L$ and $\mathcal{I}_L$ do not include the following functions
\begin{align*}
    f_1(x)=\frac{1-L}{1+L}x\,\,\,\,\,\,\text{and}\,\,\,\,\,\,f_2(x)=\frac{1+L}{1-L}x,
\end{align*}
which are in $\mathcal{M}_L$, since
\begin{align*}
    \mathrm{Lip}\left(\frac{1-L^2}{1+L^2}f_1 - \mathrm{Id}\right) = \mathrm{Lip}\left(\frac{1-L^2}{1+L^2}f_2 - \mathrm{Id}\right) = \frac{2L}{1+L^2}.
\end{align*}
Hence (iii) and (iv) hold. 
See Figure \ref{fig:compare_RLILML} in the main text for visualization, where $p$ is fixed as a unit vector pointing in the $+y$ direction. \qed

\subsection{Equivalence under the limit \texorpdfstring{$L\rightarrow1^-$}{L->1-}}

For each function class, if we consider the union of the ``subscript $L$-sets'' for all $0\le L < 1$, then we have the following theorem.
\begin{thm}\label{thm:rep_equivalence}
	Define the set $A$ as
	\begin{align*}
		A := \left\lbrace F\in C^2(\mathbb{R}^n,\mathbb{R}^n) \bigg| \text{$F$ is $\eta$-strongly monotone and $\nu$-Lipschitz for some $\eta,\nu > 0$} \right\rbrace.
	\end{align*}
	Then, $\mathbb{R}^+\mathcal{R}_{\lbrack0,1)} = \mathbb{R}^+\mathcal{I}_{\lbrack0,1)} = \mathbb{R}^+\mathcal{M}_{\lbrack0,1)} = A$. Here, $\mathcal{R}_{\lbrack0,1)} = \cup_{L\in\lbrack0,1)} \mathcal{R}_L$, and the same for $\mathcal{I}_{\lbrack0,1)}$ and $\mathcal{M}_{\lbrack0,1)}$.
\end{thm}

\textit{Proof.} The equivalence between $\mathbb{R}^+\mathcal{R}_{\lbrack0,1)}$ and $\mathbb{R}^+\mathcal{I}_{\lbrack0,1)}$ directly follows from Theorem~\ref{thm:relationship_between_function_classes}. The equivalence between $\mathbb{R}^+\mathcal{R}_{\lbrack0,1)}$ and $\mathbb{R}^+\mathcal{M}_{\lbrack0,1)}$ also follows from Theorem~\ref{thm:relationship_between_function_classes} by noting that the function $f(t) = 2t/(1+t^2)$ is a bijection from $\lbrack 0,1)$ to itself.

It remains to prove the last equality. First, $\mathbb{R}^+\mathcal{M}_{\lbrack0,1)}\subseteq A$ holds because each function in $\mathcal{M}_L$ is $(1-L)/(1+L)$-strongly monotone and $(1+L)/(1-L)$-Lipschitz for all $0\le L < 1$, by Theorem~\ref{thm:monotone_op_L_Lipschitz_Cayley}.

We now prove the reverse direction. For any $F\in A$, there exists a $0 < K \le 1$ such that $K \le \eta$ and $\nu \le 1/K$, so that $F$ is $K$-strongly monotone and $1/K$-Lipschitz. For any $x_1,x_2\in\mathbb{R}^n$, write $p = x_1-x_2$ and $q = F(x_1) - F(x_2)$ as before. We have
\begin{align}\label{eq:K_strongly_monotone_1K_lipschitz}
	\lVert q\rVert_2 \le \frac{1}{K}\lVert p\rVert_2, \,\,\,\, \langle q,p \rangle \ge K\lVert p\rVert_2^2.
\end{align}
Now, $F\in\mathcal{M}_L$ for some $0\le L < 1$ if and only if
\begin{align}\label{eq:monotone_L_lipschitz}
	\lVert q-p\rVert_2 \le L\lVert q+p\rVert_2
\end{align} for all $x_1, x_2\in\mathbb{R}^n$. To find a sufficient condition for $L$, we manipulate this equation:
\begin{align*}
	&\lVert q-p\rVert_2 \le L\lVert q+p\rVert_2 \\
	&\Leftrightarrow 2(L^2+1)\langle q,p\rangle - (1-L^2)(\lVert p\rVert_2^2 + \lVert q\rVert_2^2) \ge 0.
\end{align*}
Since
\begin{align*}
	2(L^2+1)\langle q,p\rangle - (1-L^2)(\lVert p\rVert_2^2 + \lVert q\rVert_2^2) &\ge \left\lbrack 2(L^2+1) K - (1-L^2)\left(1 + \frac{1}{K^2}\right) \right\rbrack \lVert p\rVert_2^2,
\end{align*}
by (\ref{eq:K_strongly_monotone_1K_lipschitz}), the inequality (\ref{eq:monotone_L_lipschitz}) will hold if we can find a $0\le L < 1$ such that
\begin{align*}
	2(L^2+1) K - (1-L^2)\left(1 + \frac{1}{K^2}\right) \ge 0,
\end{align*}
which is indeed solved by any choice of $L$ satisfying
\begin{align*}
	\sqrt{\frac{1+K^2-2K^3}{1+K^2+2K^3}} \le L < 1.
\end{align*}
Since
\begin{align*}
	\sqrt{\frac{1+K^2-2K^3}{1+K^2+2K^3}} < 1,
\end{align*}
for all $0 < K \le 1$, there always exists such an $L$. This proves $A\subseteq \mathbb{R}^+\mathcal{M}_{\lbrack0,1)}$. Hence, $\mathbb{R}^+\mathcal{M}_{\lbrack0,1)} = A$. \qed

\subsection{The properties of monotone operators built from \texorpdfstring{$L$}{L}-Lipschitz operators}\label{sec:monotone_op_L_Lipschitz_Cayley}

\begin{thm}\label{thm:monotone_op_L_Lipschitz_Cayley}
    Let $G$ be an $L$-Lipschitz operator for $L<1$. The monotone operator $F$ having $G$ as its Cayley operator is (i) $\eta$-strongly monotone and (ii) $\nu$-Lipschitz for
    \begin{align*}
        \eta=\frac{1-L}{1+L}\,\,\,\,\,\,\text{and}\,\,\,\,\,\,\nu=\frac{1+L}{1-L}.
    \end{align*}
\end{thm}

\textit{Proof.} Let $(x_1,y_1),\,(x_2,y_2)\in F$ and $a_i = x_i+y_i$, $b_i=x_i-y_i$ for $i=1,2$. We first note that $\lVert b_1-b_2\rVert_2 = \lVert G(a_1)-G(a_2)\rVert_2\le L\lVert a_1-a_2\rVert_2$ by the definition of Lipschitz continuity.

(i) Let's prove the $\eta$-strongly monotonicity. We have
\begin{align*}
    \left\langle y_1-y_2,x_1-x_2\right\rangle &= \left\langle \frac{1}{2}(a_1-b_1)-\frac{1}{2}(a_2-b_2), \frac{1}{2}(a_1+b_1)-\frac{1}{2}(a_2+b_2)\right\rangle \\
    &= \frac{1}{4}\left( \lVert a_1-a_2\rVert_2^2 - \lVert b_1-b_2\rVert_2^2 \right)\ge \frac{1-L^2}{4}\lVert a_1-a_2\rVert_2^2,
\end{align*}
and
\begin{align*}
    \lVert x_1-x_2\rVert_2 &= \left\lVert \frac{1}{2}(a_1+b_1) - \frac{1}{2}(a_2+b_2)\right\rVert_2 \\
    &\le \frac{1}{2}\left( \lVert a_1-a_2\rVert_2 + \lVert b_1-b_2\rVert_2 \right) \le \frac{1+L}{2}\lVert a_1-a_2\rVert_2.
\end{align*}
Combining the two inequalities we have
\begin{align*}
    \langle y_1-y_2,x_1-x_2\rangle \ge \frac{1}{4}(1-L^2)\left(\frac{2}{1+L}\lVert x_1-x_2\rVert_2\right)^2 = \frac{1-L}{1+L} \lVert x_1-x_2\rVert_2^2,
\end{align*}
which implies the $\eta$-strong monotonicity. Equality holds with $G(x) = Lx$.

(ii) Let's prove the $\nu$-Lipschitzness. We have
\begin{align*}
    \lVert y_1-y_2\rVert_2 &= \left\lVert \frac{1}{2}(a_1-b_1)-\frac{1}{2}(a_2-b_2)\right\rVert_2 \\
    &\le \frac{1}{2}\left( \lVert a_1-a_2\rVert_2 + \lVert b_1-b_2\rVert_2 \right) \le \frac{1+L}{2} \lVert a_1-a_2\rVert_2,
\end{align*}
and
\begin{align*}
    \lVert x_1-x_2\rVert_2 &= \left\lVert \frac{1}{2}(a_1+b_1)-\frac{1}{2}(a_2+b_2)\right\rVert_2 \\
    &\ge \frac{1}{2}\left(\lVert a_1-a_2\rVert_2-\lVert b_1-b_2\rVert_2\right) \ge \frac{1-L}{2}\lVert a_1-a_2\rVert_2.
\end{align*}
Combining the two inequalities we have
\begin{align*}
	\left\lVert y_1-y_2 \right\rVert_2 \le \frac{1+L}{2}\left\lVert a_1-a_2\right\rVert_2 \le \frac{1+L}{1-L} \left\lVert x_1-x_2 \right\rVert_2,
\end{align*}
which implies the $\nu$-Lipschitzness. Equality holds with $G(x) = -Lx$. \qed

\clearpage

\section{Computation}
    \subsection{Forward and backward algorithms for a single monotone layer}

We describe the forward and backward algorithms for a single monotone layer. Both utilize fixed-point iterations, but the forward pass uses non-linear functions, whereas the backward pass uses linear functions. We let $G_\theta$ denote the $G$-network in Definition~\ref{defn:monotone_flow} with parameters $\theta$, $F_\theta$ the monotone formulation of $G_\theta$, and $\ell$ the loss function of the whole model. Note that in Algorithm~\ref{alg:fixed_point_solver}, we maintain the values of $\alpha$ and $\beta$ for each sample in minibatch. Also note that in Algorithm~\ref{alg:forward}, the maximum number of iterations $n_\mathrm{max}$ applies separately to the first and the second fixed-point algorithm. We choose $(\epsilon,\,n_{\mathrm{max}})=(10^{-6}, 2000)$ for the forward pass and $(\epsilon,\,n_{\mathrm{max}})=(10^{-9}, 100)$ for the backward pass.

For time and memory efficiency, we use the Neumann gradient estimator following~\cite{ResidualFlows}, defined by the following equation. The last expression has the form $\partial\mathcal{L}/\partial\theta$; we use $\mathcal{L}$ as the \emph{surrogate loss} for estimating the gradients with respect to $\theta$.
\begin{equation}\label{eq:neumann_grad_estimator}
    \begin{aligned}
        \frac{\partial}{\partial\theta}\log\det J_{F_\theta}
        &= \frac{\partial}{\partial\theta}\mathrm{tr}\left\lbrack \log(I-J_{G_\theta}) - \log(I+J_{G_\theta})\right\rbrack \\
        &= \frac{\partial}{\partial\theta}\mathrm{tr}\left\lbrack \sum_{k=1}^{\infty} \frac{(-1) - (-1)^{k+1}}{k}J_{G_\theta}^k\right\rbrack \\
        &= \mathrm{tr}\left\lbrack \sum_{k=1}^{\infty} ((-1)-(-1)^{k+1})\frac{\partial J_{G_\theta}}{\partial\theta}J_{G_\theta}^{k-1}\right\rbrack \\
        &= \mathbb{E}_{n\sim p_N(n),v\sim\mathcal{N}(0,I)}\left\lbrack \sum_{k=1}^{n} \frac{(-1)-(-1)^{k+1}}{P(N\ge k)} v^T \frac{\partial J_{G_\theta}}{\partial\theta} J_{G_\theta}^{k-1} v \right\rbrack \\
        &= \frac{\partial}{\partial\theta}\mathbb{E}_{n\sim p_N(n),v\sim\mathcal{N}(0,I)}\left\lbrack \sum_{k=1}^{n} \frac{(-1)-(-1)^{k+1}}{P(N\ge k)} v^T J_{G_\theta} \mathrm{stop\_gradient}\left(J_{G_\theta}^{k-1} v\right) \right\rbrack.
    \end{aligned}
\end{equation}

\newcommand{\var}[1]{{\ttfamily#1}}

\begin{algorithm}[h]
    \caption{Forward algorithm for a single monotone layer}\label{alg:forward}
    \begin{algorithmic}[1]
        \State \textbf{Require:} $G_\theta$, tolerance $\epsilon$, max. number of iterations $n_{\mathrm{max}}$
        \State \textbf{Input:} $x$
        \State \textbf{Output:} $z=F_\theta(x)$, $\displaystyle p = \log\det\left|\frac{\partial z}{\partial x}\right|$
        \State $u \leftarrow 2x$
        \State $w \leftarrow $ FixedPointSolver$\left(f(y) = u - G_\theta(y),\, x,\, \epsilon,\, n_{\mathrm{max}} \right)$
        \State $z \leftarrow w - x$
        \If {training}
            \State Estimate $p$ using the surrogate loss~\eqref{eq:neumann_grad_estimator}.
        \Else
            \State Estimate $p$ using the true estimator~\eqref{eq:monotone_flow_logdet}.
        \EndIf
        \State Return $z$ and $p$
    \end{algorithmic}
\end{algorithm}

\begin{algorithm}[ht]
    \caption{Backward algorithm for a single monotone layer}\label{alg:backward}
    \begin{algorithmic}[1]
        \State \textbf{Require:} $G_\theta$, tolerance $\epsilon$, max. number of iterations $n_{\mathrm{max}}$
        \State \textbf{Input:} $x$, $z$, $p$, $\displaystyle \frac{\partial\ell}{\partial z}$, $\displaystyle \frac{\partial\ell}{\partial p}$
        \State \textbf{Output:} $\displaystyle \frac{\partial\ell}{\partial x}$, $\displaystyle \frac{\partial\ell}{\partial\theta}$
        \State Backpropagate through $z$ to $w$ and $x$
        \State Backpropagate through $w$ to $u$ and $\theta$:
        \State \hspace{0.5cm} $g \leftarrow$ FixedPointSolver$\displaystyle \left( f(y) = \frac{\partial\ell}{\partial w} - y\frac{\partial G_\theta(w)}{\partial w},\, \frac{\partial\ell}{\partial w},\, \epsilon,\, n_{\mathrm{max}} \right)$
        \State \hspace{0.5cm} $\displaystyle \frac{\partial\ell}{\partial u} \leftarrow g$
        \State \hspace{0.5cm} $\displaystyle \frac{\partial\ell}{\partial\theta} \leftarrow -g\frac{\partial G_\theta(w)}{\partial\theta}$
        \State Backpropagate through $u$ to $x$
        \State Backpropagate through $p$ to $x$ and $\theta$
        \State Return the gradients $\displaystyle \frac{\partial\ell}{\partial x}$ and $\displaystyle \frac{\partial\ell}{\partial\theta}$
    \end{algorithmic}
\end{algorithm}

\begin{algorithm}[ht]
    \caption{Fixed-point solver}\label{alg:fixed_point_solver}
    \begin{algorithmic}[1]
        \State \textbf{Require:} input function (contraction mapping) $f$, tolerance $\epsilon$, max. number of iterations $n_{\mathrm{max}}$
        \State \textbf{Input:} $y_0$, the starting point of the fixed-point iteration
        \State \textbf{Output:} $y$, the fixed point of $f$ satisfying $y=f(y)$
        \State $y_1 \leftarrow f(y_0)$
        \State $f_\mathrm{prev} \leftarrow y_1$
        \State $\mathrm{d}y_\mathrm{prev} \leftarrow y_1 - y_0$
        \State $y_\mathrm{curr} \leftarrow y_1$
        \While {error higher than the tolerance $\epsilon$ and iteration limit $n_{\mathrm{max}}$ not reached}
            \If {error has not improved in 10 recent iterations}
                \State \textbf{break}
            \EndIf
            \State $f_\mathrm{curr} \leftarrow f(y_\mathrm{curr})$
            \State $\mathrm{d}y_\mathrm{curr} \leftarrow f_\mathrm{curr} - y_\mathrm{curr}$
            \State $\mathrm{d}^2y_\mathrm{curr} \leftarrow \mathrm{d}y_\mathrm{curr} - \mathrm{d}y_\mathrm{prev}$
            \State $\displaystyle \beta \leftarrow \frac{\langle \mathrm{d}^2y_\mathrm{curr}, \mathrm{d}y_\mathrm{curr} \rangle}{||\mathrm{d}^2y_\mathrm{curr}||_2^2 + 10^{-8}}$
            \State $y_\mathrm{next} \leftarrow f_\mathrm{curr} - \beta(f_\mathrm{curr}-f_\mathrm{prev})$
            \State $\mathrm{d}y_\mathrm{prev} \leftarrow \mathrm{d}y_\mathrm{curr}$
            \State $f_\mathrm{prev} \leftarrow f_\mathrm{curr}$
            \State $y_\mathrm{curr} \leftarrow y_\mathrm{next}$
        \EndWhile
        \If {error within the tolerance $\epsilon$}
            \State Return $y_\mathrm{curr}$
        \EndIf
        \State $\alpha \leftarrow 0.5$
        \While {error higher than the tolerance $\epsilon$ and iteration limit $n_{\mathrm{max}}$ not reached}
            \If {error has not improved in recent 30 iterations}
                \State $\alpha \leftarrow \max\left\lbrace0.9\alpha, 0.1\right\rbrace$
                \State Reset the counter
            \EndIf
            \State $y_\mathrm{curr}\leftarrow (1-\alpha)y_\mathrm{curr} + \alpha f(y_\mathrm{curr})$
        \EndWhile
        \State Print the current error for logging purposes
        \State Return $y_\mathrm{curr}$
    \end{algorithmic}
\end{algorithm}

\clearpage

\section{Experimental details}
    All of our experiments are implemented with PyTorch, based on the public code of i-DenseNets~\cite{i-DenseNets}. We have implemented CPila as a C++ CUDA extension for speedup. In all experiments, we disable the TensorFloat32 (TF32) and the CUDA benchmarks option to prevent the non-deterministic computation from affecting the convergence of the fixed-point iterations.

\subsection{1D toy experiment}\label{apn:experiment_toy_1d}

\textbf{Training data.} We fit a strongly monotone function $s: \mathbb{R}\rightarrow\mathbb{R}$ defined as follows:
\begin{equation}
    s(x) = t(x+2) + t(x+1) + t(x) + t(x-1),
\end{equation}
where $t: \mathbb{R}\rightarrow\mathbb{R}$ is defined as
\begin{equation}
    t(x) = \begin{cases}
        0, \hspace{4.0cm} x < 0 \\
        \displaystyle \max\left\lbrace \alpha x, \frac{1}{\alpha}(x-1+\alpha) \right\rbrace, \hspace{0.25cm} 0 \le x \le 1 \\
        1, \hspace{4.0cm} x > 1
    \end{cases}
\end{equation}
with the choice of $\alpha=0.05$. The function $s$ is considered on the interval $\lbrack-2,2\rbrack$ only.

\textbf{Training objective.} We use the mean squared error (MSE) for the loss function and report the best test MSE in Figure~\ref{fig:toy_1d}.

\textbf{Training procedure.} We use Adam~\cite{Adam} for optimization with $(\beta_1,\,\beta_2)=(0.9,\,0.99)$, eps = $10^{-8}$, and an initial learning rate of 0.01, which drops to 0.002 and 0.0004 after 5,000 and 10,000 iterations, respectively. We train for 15,000 iterations in total. We do not apply weight decay. We enable the learnable concatenation at the beginning. For training, we randomly sample 5,000 points uniformly from $\lbrack-2, 2\rbrack$ on each iteration; for testing, we use 20,001 uniformly spaced points on $\lbrack-2,2\rbrack$. We run tests every 100 iterations. We train each of the four models using a single NVIDIA RTX 2080 Ti.

\textbf{Model architecture.} There are four building blocks: ResidualBlock (RB), InverseResidualBlock (IRB), MonotoneBlock (MB), and ActNorm (learnable scaling; LS). RB, IRB, and MB share the same $G$-network built by four DenseNet concatenations, each adding 128 channels, followed by a final linear layer. We use Concatenated ReLU (CReLU) as the activation function. The Lipschitz constant for spectral normalization and dense layers are 0.99 and 0.99, respectively; the tolerance for spectral normalization is $10^{-4}$. ActNorm~\cite{GLOW} learns a positive scaling parameter and a bias for each channel. It acts as a learnable scaling (LS) layer that applies to the whole input since there is only one channel for 1D toy experiments. The architectures for each model are as follows:

\begin{itemize}
    \item (a) 2 RBs w/o LS:
    
    \centerline{ResidualBlock $\rightarrow$ ResidualBlock}
    \item (b) 2 RBs w/ LS:
    
    \centerline{ActNorm $\rightarrow$ ResidualBlock $\rightarrow$ ActNorm $\rightarrow$ ResidualBlock $\rightarrow$ ActNorm}
    \item (c) 1 RB + 1 IRB w/ LS:

    \centerline{ActNorm $\rightarrow$ ResidualBlock $\rightarrow$ ActNorm $\rightarrow$ InverseResidualBlock $\rightarrow$ ActNorm}
    \item (d) 2 MBs w/ LS:
    
    \centerline{ActNorm $\rightarrow$ MonotoneBlock $\rightarrow$ ActNorm $\rightarrow$ MonotoneBlock $\rightarrow$ ActNorm}
\end{itemize}

\subsection{2D toy experiments}\label{apn:experiment_toy_2d}

\textbf{Training data.} We use the eight 2D toy densities provided with the source code of i-DenseNets~\cite{i-DenseNets}, which originate from~\cite{FFJORD} to the best of our knowledge. They consist of `2 Spirals', `8 Gaussians', `Checkerboard', `Circles', `Moons', `Pinwheel', `Rings', and `Swissroll'.

\textbf{Training objective.} We use the typical log-likelihood objective of normalizing flows. Note that we evaluate the Jacobian determinant exactly using the determinant formula for 2D matrices since the low dimensionality (2D) allows for the fast and exact computation of the Jacobian determinant without needing to resort to the stochastic estimation.

\textbf{Training procedure.} We use Adam~\cite{Adam} for optimization with $(\beta_1,\,\beta_2)=(0.9,\,0.99)$, eps = $10^{-8}$, and the learning rate of 0.001. We train for 50,000 iterations in total. We apply the weight decay of $10^{-5}$. We enable the learnable concatenation after 25,000 iterations. For training, we sample 500 points from the target distribution on each iteration; for testing, we sample 10,000 points. We run tests every 100 iterations. We train i-DenseNets and Monotone Flows for each dataset using a single NVIDIA RTX 2080 Ti.

\textbf{Model architecture.} The $G$-networks for i-DenseNets and Monotone Flows only differ in the choice of the activation function, where we use CLipSwish and CPila, respectively. Otherwise, they share the same structure, built by three DenseNet concatenations, each adding 16 channels, followed by a final linear layer. The Lipschitz constant for spectral normalization and dense layers are 0.90 and 0.98, respectively. We do not explicitly specify the tolerance for spectral normalization; instead, we run five iterations on each spectral normalization update regardless of the current error. We do not use ActNorm layers for 2D toy experiments. The architectures for each model are as follows; we use ten blocks following the 2D toy experiment setup of~\cite{i-DenseNets}.

\begin{itemize}
    \item i-DenseNets: 10 $\times$ [ResidualBlock]
    \item Monotone Flows: 10 $\times$ [MonotoneBlock]
\end{itemize}

\subsection{Image experiments}\label{apn:experiment_images}

\textbf{Training data.} We use MNIST, CIFAR-10, ImageNet32, and ImageNet64. For CIFAR-10, we apply a random horizontal flip with a probability of 0.5 and do nothing for the other three datasets. We then add noise for uniform dequantization. The images keep their original sizes, which are $28\times28$, $32\times32$, $32\times32$, and $64\times64$, respectively.

\textbf{Training objective.} We use the typical log-likelihood objective of normalizing flows. We evaluate the Jacobian determinant using the stochastic estimator, with ten exact terms plus additional terms with the Russian roulette estimator. In contrast to the 2D toy experiments, we use the bits per dimension, defined as the $\log_2$-likelihood divided by the input dimension, with the compensation for the quantization levels. Since there are 256 quantization levels for the image data, we add 8 to the $\log_2$-likelihood.

\textbf{Training procedure.} We use Adam~\cite{Adam} for optimization with $(\beta_1,\,\beta_2)=(0.9,\,0.99)$ and eps = $10^{-8}$. We do not apply weight decay. Other options vary among the four datasets, as described in Table~\ref{tab:image_experimental_setup_training}. We run tests at the end of every epoch. We use four RTX 3090 GPUs for MNIST and CIFAR-10 and eight A100 GPUs for ImageNet32 and ImageNet64.

\begin{table}[t]
    \centering
    \captionsetup{width=\textwidth}
    \caption{The training setup for image density estimation tasks. LC: learnable concatenation.}
    \label{tab:image_experimental_setup_training}
    \begin{tabular}{lcccc}
        \hline
        & MNIST & CIFAR-10 & ImageNet32 & ImageNet64 \\
        \hline
        Learning rate & 0.001 & 0.001 & 0.004 & 0.004 \\
        Batch size & 64 & 64 & 256 & 256 \\
        Number of epochs & 100 & 1,000 & 20 & 20 \\
        Enable LC & After 25 epochs & After 25 epochs & At beginning & At beginning \\
        \hline
    \end{tabular}
\end{table}

\textbf{Model architecture.} We use monotone blocks for each model, with the $G$-networks built by three DenseNet concatenations, each adding 172 channels, followed by a final linear layer. We use Concatenated Pila (CPila) as the activation function. The Lipschitz constant for spectral normalization and dense layers are 0.98 and 0.98, respectively. We set the tolerance for spectral normalization to $10^{-3}$. To implement a multi-scale architecture, we utilize invertible downsampling operations (`Squeeze'), which package the pixels of two-dimensional $2\times2$ cells into a 4-dimensional vector, hence converting an input of the shape $C\times H\times W$ into the shape $4C\times \frac{H}{2}\times \frac{W}{2}$. Layers with no invertible downscaling layers in between share the same scale. We use four fully-connected layers at the end of the models with a DenseNet depth and growth of 3 and 16. Note that the fully-connected layers first project the input into a vector of 64 dimensions, apply the DenseNet formulation, and then project back to the original dimensions. `FactorOut' is the operation that factors out half of the variables at the transition of one scale to another scale, splitting across channels. `Squeeze2d' and `FactorOut' applied in succession converts an input of the shape $C\times H\times W$ into the shape $2C\times \frac{H}{2}\times \frac{W}{2}$. The architectural choices that differ across the datasets are displayed in Table~\ref{tab:image_experimental_setup_model}. The architectures for each model are as follows:

\begin{table}[t]
    \centering
    \captionsetup{width=\textwidth}
    \caption{The model setup for image density estimation tasks.}
    \label{tab:image_experimental_setup_model}
    \begin{tabular}{lcccc}
        \hline
        & MNIST & CIFAR-10 & ImageNet32 & ImageNet64 \\
        \hline
        Logit transform's $\alpha$ & $10^{-6}$ & 0.05 & 0.05 & 0.05 \\
        Number of scales & 3 & 3 & 3 & 3 \\
        Number of flow blocks per scale & 16 & 16 & 32 & 32 \\
        Factor out at the end of each scale & No & No & No & Yes \\
        \hline
    \end{tabular}
\end{table}

The architectures for each model are as follows. Conv: convolutional layers; FC: fully connected layers.

\begin{itemize}
    \item MNIST: LogitTransform($10^{-6}$) $\rightarrow$ ActNorm2d $\rightarrow$ 16 $\times$ [MonotoneBlock$_\mathrm{Conv}$ $\rightarrow$ ActNorm2d] $\rightarrow$ Squeeze2d $\rightarrow$ 16 $\times$ [MonotoneBlock$_\mathrm{Conv}$ $\rightarrow$ ActNorm2d] $\rightarrow$ Squeeze2d $\rightarrow$ 15 $\times$ [MonotoneBlock$_\mathrm{Conv}$ $\rightarrow$ ActNorm2d] $\rightarrow$ MonotoneBlock$_\mathrm{Conv}$ $\rightarrow$ ActNorm1d $\rightarrow$ 4 $\times$ [MonotoneBlock$_\mathrm{FC}$ $\rightarrow$ ActNorm1d]
    
    \item CIFAR-10: LogitTransform($0.05$) $\rightarrow$ ActNorm2d $\rightarrow$ 16 $\times$ [MonotoneBlock$_\mathrm{Conv}$ $\rightarrow$ ActNorm2d] $\rightarrow$ Squeeze2d $\rightarrow$ 16 $\times$ [MonotoneBlock$_\mathrm{Conv}$ $\rightarrow$ ActNorm2d] $\rightarrow$ Squeeze2d $\rightarrow$ 15 $\times$ [MonotoneBlock$_\mathrm{Conv}$ $\rightarrow$ ActNorm2d] $\rightarrow$ MonotoneBlock$_\mathrm{Conv}$ $\rightarrow$ ActNorm1d $\rightarrow$ 4 $\times$ [MonotoneBlock$_\mathrm{FC}$ $\rightarrow$ ActNorm1d]
    
    \item ImageNet32: LogitTransform($0.05$) $\rightarrow$ ActNorm2d $\rightarrow$ 32 $\times$ [MonotoneBlock$_\mathrm{Conv}$ $\rightarrow$ ActNorm2d] $\rightarrow$ Squeeze2d $\rightarrow$ 32 $\times$ [MonotoneBlock$_\mathrm{Conv}$ $\rightarrow$ ActNorm2d] $\rightarrow$ Squeeze2d $\rightarrow$ 31 $\times$ [MonotoneBlock$_\mathrm{Conv}$ $\rightarrow$ ActNorm2d] $\rightarrow$ MonotoneBlock$_\mathrm{Conv}$ $\rightarrow$ ActNorm1d $\rightarrow$ 4 $\times$ [MonotoneBlock$_\mathrm{FC}$ $\rightarrow$ ActNorm1d]
    
    \item ImageNet64: LogitTransform($0.05$) $\rightarrow$ ActNorm2d $\rightarrow$ 32 $\times$ [MonotoneBlock$_\mathrm{Conv}$ $\rightarrow$ ActNorm2d] $\rightarrow$ Squeeze2d $\rightarrow$ FactorOut $\rightarrow$ 32 $\times$ [MonotoneBlock$_\mathrm{Conv}$ $\rightarrow$ ActNorm2d] $\rightarrow$ Squeeze2d $\rightarrow$ FactorOut $\rightarrow$ 31 $\times$ [MonotoneBlock$_\mathrm{Conv}$ $\rightarrow$ ActNorm2d] $\rightarrow$ MonotoneBlock$_\mathrm{Conv}$ $\rightarrow$ ActNorm1d $\rightarrow$ 4 $\times$ [MonotoneBlock$_\mathrm{FC}$ $\rightarrow$ ActNorm1d]
\end{itemize}

\subsection{Variational dequantization}\label{apn:experiment_vardeq}

The experiments with variational dequantization use the same main network structure used for the experiments with uniform dequantization. The variational dequantization network, now added at the beginning of the network, uses three invertible DenseNet blocks, which uses CPila activation and does not use monotone formulation. Conditioning is performed by injecting conditional feature vectors into each step inside each invertible DenseNet block. The conditional feature vectors are computed from the input image using ResNet-based neural networks.

\subsection{Ablation studies}\label{apn:experiment_abl}

The ablation studies are mostly the same as the image experiments discussed in Appendix~\ref{apn:experiment_images}. Thus, we only highlight the differences:
\begin{itemize}
    \item When ablating the monotone formulation, we replace each MonotoneBlock$_\mathrm{Conv/FC}$ with ResidualBlock$_\mathrm{Conv/FC}$, while keeping the $G$-networks the same.
    \item When ablating the CPila activation function, we replace all occurrences of CPila in each MonotoneBlock$_\mathrm{Conv/FC}$ or  ResidualBlock$_\mathrm{Conv/FC}$ with CLipSwish.
\end{itemize}

\subsection{Classification experiments}\label{apn:experiment_classification}

To verify the capacity of Monotone Flows, we perform classification experiments with the same network used for image density estimation tasks.

\textbf{Training data.} We use the CIFAR-10 dataset. We apply the standard data augmentation for CIFAR-10, which amounts to padding the input image by 4 pixels, randomly cropping the image back to the original size, and applying a random horizontal flip. Then, we add noise following the image density estimation experiments.

\textbf{Training objective.} We use the cross-entropy loss during training and report the average test accuracy over the last five epochs.

\textbf{Training procedure.} We use Adam~\cite{Adam} for optimization with $(\beta_1,\,\beta_2)=(0.9,\,0.99)$, eps = $10^{-8}$, and the learning rate of 0.001. We do not apply weight decay. We enable the learnable concatenation after 25 epochs. We train for 200 epochs with a batch size of 128 and run tests at the end of every epoch. We train each model using two NVIDIA V100 GPUs.

\textbf{Model architecture.} Each model has the following structure: Mean-Std normalization $\rightarrow$ $k$ $\times$ [MonotoneBlock$_\mathrm{Conv}$ $\rightarrow$ ActNorm2d] $\rightarrow$ Squeeze2d $\rightarrow$ $k$ $\times$ [MonotoneBlock$_\mathrm{Conv}$ $\rightarrow$ ActNorm2d] $\rightarrow$ Squeeze2d $\rightarrow$ $k$ $\times$ [MonotoneBlock$_\mathrm{Conv}$ $\rightarrow$ ActNorm2d]. The $G$-network has a DenseNet depth and growth of 3 and 80. There are no fully-connected layers at the end. We use $k=1$ for tiny models, $k=4$ for small models, and $k=16$ for large models. For i-DenseNets, each MonotoneBlock gets replaced with a ResidualBlock with the CLipSwish activation function. Classification heads are attached after each Squeeze2d and at the end of the model. Each head consists of Conv2d $\rightarrow$ ActNorm2d $\rightarrow$ ReLU $\rightarrow$ AvgPool2d, yielding a 256-dimensional feature vector per head. The vectors are concatenated into a 768-dimensional vector, passed through a linear layer, and then fed into a softmax layer for classification.

\textbf{Classification results.} We present the results in Table~\ref{tab:experiment_cifar10_classification}. The results clearly demonstrate that Monotone Flows consistently outperform i-DenseNets for all model sizes considered.

\begin{table}[t]
    \centering
    \caption{Classification results on CIFAR-10.}
    \label{tab:experiment_cifar10_classification}
    \begin{tabular}{lcc}
        \hline
         & i-DenseNet & Monotone Flow \\
         \hline
         Tiny $(k=1)$ & 86.7 \% & \textbf{88.9 \%} \\
         Small $(k=4)$ & 90.2 \% & \textbf{91.8 \%} \\
         Large $(k=16)$ & 92.5 \% & \textbf{93.4 \%} \\
         \hline
    \end{tabular}
\end{table}

\subsection{Training curve for CIFAR-10 density estimation}\label{apn:training_curve_CIFAR10_density}

\begin{figure}[H]
	\centering
	\begin{tikzpicture}
		\begin{axis}[
			ymin=3.20, ymax=3.30, ytick={3.20,3.22,3.24,3.26,3.28,3.30}, y=30.0cm,
			xmin=0, xmax=1000, xtick={0,200,400,600,800,1000}, x=0.012cm,
			xlabel={Epochs},
			ylabel={Test bits/dim},
		    no markers,
            every axis plot/.append style={thin}
			]
			\addplot+ [cyan] table [x=epoch, y=id, col sep=comma, sharp plot] {Figures/loss_curve.csv};
			\addplot+ [red] table [x=epoch, y=mf, col sep=comma, sharp plot] {Figures/loss_curve.csv};
			\addplot+ [orange] table [x=epoch, y=abl_mf, col sep=comma, sharp plot] {Figures/loss_curve.csv};
			\addplot+ [blue] table [x=epoch, y=abl_cpila, col sep=comma, sharp plot] {Figures/loss_curve.csv};
		\end{axis}
	\end{tikzpicture}
	\vspace{-0.3cm}
	\captionsetup{width=\linewidth}
	\caption{Loss curves for CIFAR-10 training with uniform dequantization. From top to bottom, the cyan curve denotes the baseline i-DenseNets model; the orange curve denotes our model with monotone formulation ablated; the blue curve denotes our model with CPila ablated; the red curve denotes our full model.}
\end{figure}
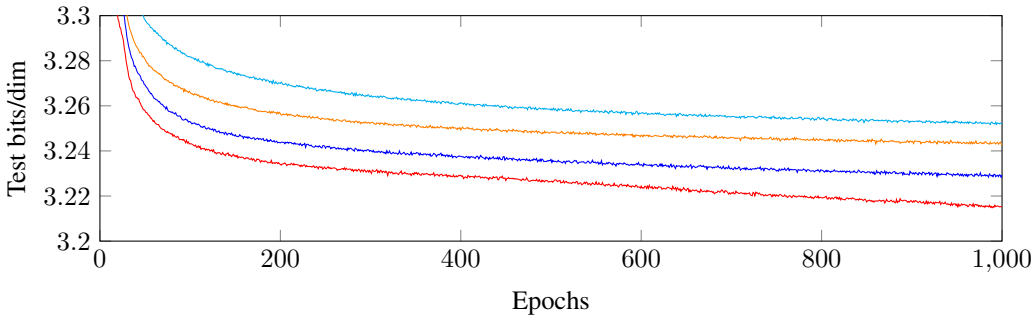

\clearpage

\section{Full toy results}
    \label{apn:toy_full_results}
\begin{figure}[H]
    \centering
    \vspace{-0.1cm}
    \begin{tabular}{cccc}
        \hline
        Dataset & Sample data & i-DenseNet & Monotone Flow \\
        \hline \\
        \vspace{-0.65cm} \\
        2 Spirals & \raisebox{-.43\height}{\includegraphics[width=25mm]{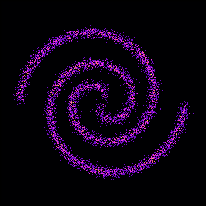}} & \raisebox{-.43\height}{\includegraphics[width=25mm]{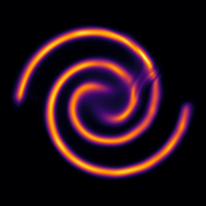}} & \raisebox{-.43\height}{\includegraphics[width=25mm]{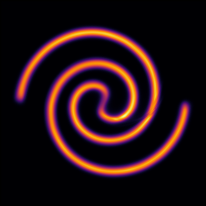}} \\
        \vspace{-0.3cm} \\
        8 Gaussians & \raisebox{-.43\height}{\includegraphics[width=25mm]{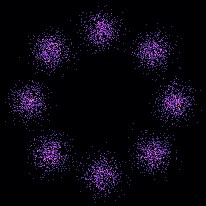}} & \raisebox{-.43\height}{\includegraphics[width=25mm]{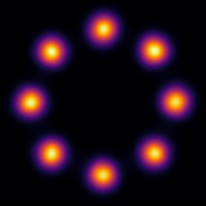}} & \raisebox{-.43\height}{\includegraphics[width=25mm]{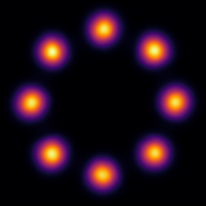}} \\
        \vspace{-0.3cm} \\
        Checkerboard & \raisebox{-.43\height}{\includegraphics[width=25mm]{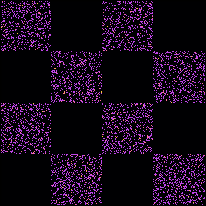}} & \raisebox{-.43\height}{\includegraphics[width=25mm]{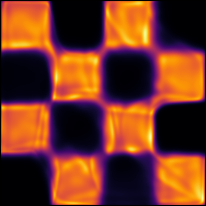}} & \raisebox{-.43\height}{\includegraphics[width=25mm]{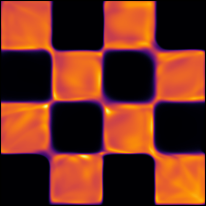}} \\
        \vspace{-0.3cm} \\
        Circles & \raisebox{-.43\height}{\includegraphics[width=25mm]{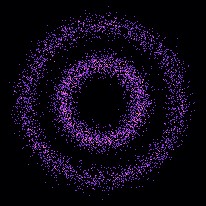}} & \raisebox{-.43\height}{\includegraphics[width=25mm]{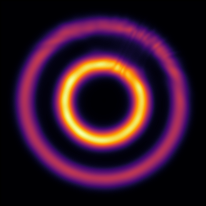}} & \raisebox{-.43\height}{\includegraphics[width=25mm]{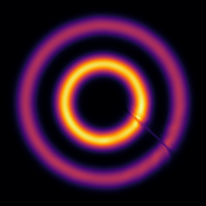}} \\
        \vspace{-0.3cm} \\
        Moons & \raisebox{-.43\height}{\includegraphics[width=25mm]{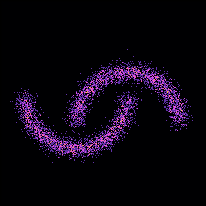}} & \raisebox{-.43\height}{\includegraphics[width=25mm]{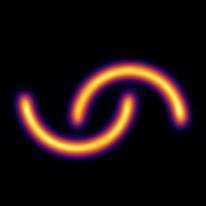}} & \raisebox{-.43\height}{\includegraphics[width=25mm]{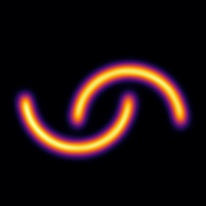}} \\
        \vspace{-0.3cm} \\
        Pinwheel & \raisebox{-.43\height}{\includegraphics[width=25mm]{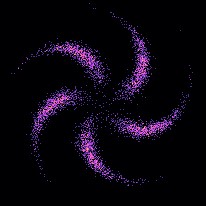}} & \raisebox{-.43\height}{\includegraphics[width=25mm]{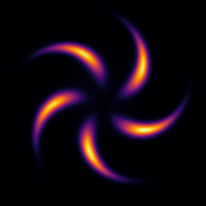}} & \raisebox{-.43\height}{\includegraphics[width=25mm]{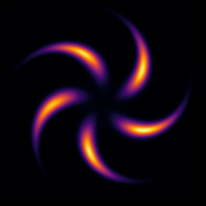}} \\
        \vspace{-0.3cm} \\
        Rings & \raisebox{-.43\height}{\includegraphics[width=25mm]{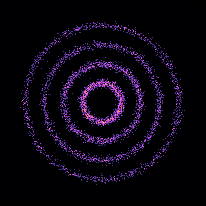}} & \raisebox{-.43\height}{\includegraphics[width=25mm]{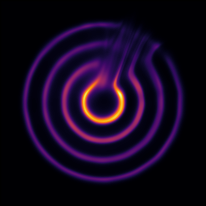}} & \raisebox{-.43\height}{\includegraphics[width=25mm]{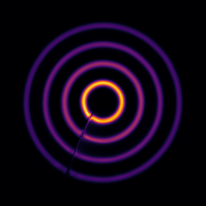}} \\
        \vspace{-0.3cm} \\
        Swissroll & \raisebox{-.43\height}{\includegraphics[width=25mm]{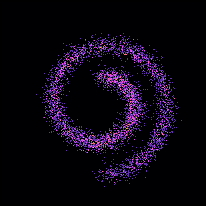}} & \raisebox{-.43\height}{\includegraphics[width=25mm]{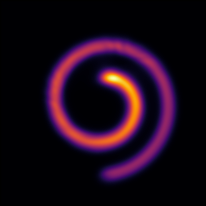}} & \raisebox{-.43\height}{\includegraphics[width=25mm]{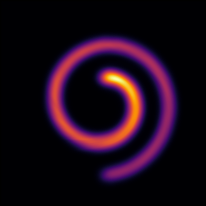}} \\
        \vspace{-0.25cm} \\
        \hline
    \end{tabular}
    \caption{Full toy results.}
\end{figure}

\clearpage

\section{Image samples}
    \label{apn:image_samples}
\begin{figure}[h]
    \centering
    \begin{subfigure}{0.49\textwidth}
        \includegraphics[width=\textwidth]{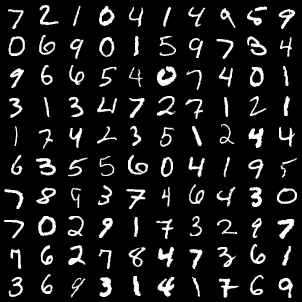}
        \caption{MNIST train data.}
    \end{subfigure}
    \hfill
    \begin{subfigure}{0.49\textwidth}
        \includegraphics[width=\textwidth]{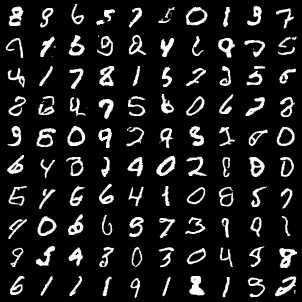}
        \caption{Monotone Flows trained on MNIST.}
    \end{subfigure}
    \caption{Train data and generated samples of MNIST.}
\end{figure}
\begin{figure}[h]
    \centering
    \begin{subfigure}{0.49\textwidth}
        \includegraphics[width=\textwidth]{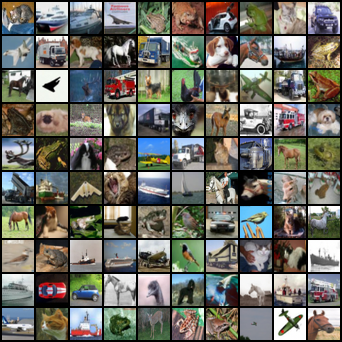}
        \caption{CIFAR-10 train data.}
    \end{subfigure}
    \hfill
    \begin{subfigure}{0.49\textwidth}
        \includegraphics[width=\textwidth]{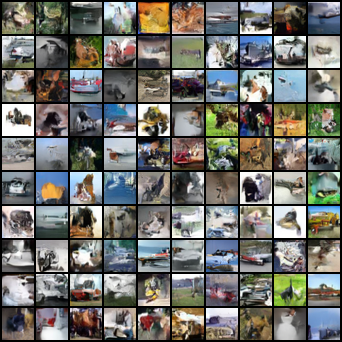}
        \caption{Monotone Flows trained on CIFAR-10.}
    \end{subfigure}
    \caption{Train data and generated samples of CIFAR-10.}
\end{figure}
\begin{figure}[t]
    \centering
    \begin{subfigure}{0.49\textwidth}
        \includegraphics[width=\textwidth]{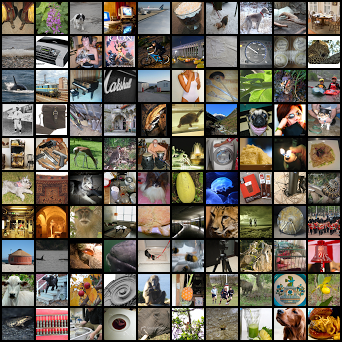}
        \caption{ImageNet32 train data.}
    \end{subfigure}
    \hfill
    \begin{subfigure}{0.49\textwidth}
        \includegraphics[width=\textwidth]{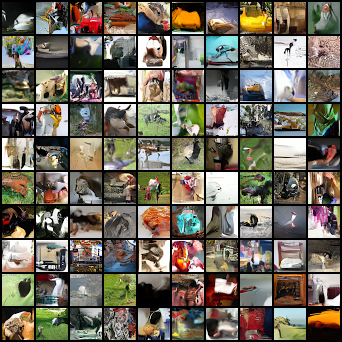}
        \caption{Monotone Flows trained on ImageNet32.}
    \end{subfigure}
    \caption{Train data and generated samples of ImageNet32.}
\end{figure}
\begin{figure}[t]
    \centering
    \begin{subfigure}{0.49\textwidth}
        \includegraphics[width=\textwidth]{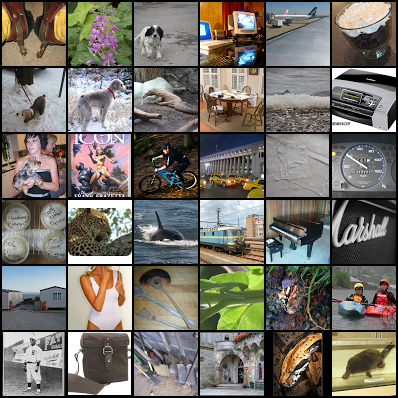}
        \caption{ImageNet64 train data.}
    \end{subfigure}
    \hfill
    \begin{subfigure}{0.49\textwidth}
        \includegraphics[width=\textwidth]{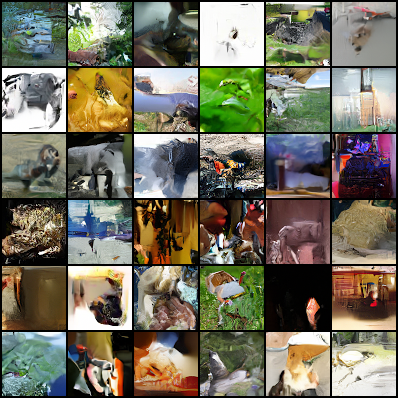}
        \caption{Monotone Flows trained on ImageNet64.}
    \end{subfigure}
    \caption{Train data and generated samples of ImageNet64.}
\end{figure}

\clearpage

\section{Limitations and negative societal impact}
    Similar to previous works: i-ResNets, Residual Flows, i-DenseNets, and Implicit Normalizing Flows, our Monotone Flows involve fixed-point equations, which often leads to computational overhead. However, the speed will likely improve as the methods for solving fixed-point equations, including neural solvers and better initialization schemes, continue to evolve. Also, although our model as a normalizing flow has the advantage of training stability and not suffering from mode collapses, the generated images generally do not yet achieve high fidelity. This is a common weakness of normalizing flow models, and we leave this for future work.

For potential negative societal impact, we note that while the improved modeling capacity of Monotone Flows can benefit many downstream applications of normalizing flows, they have the risk of being misused for the generation of fake images, just like other generative models. Hence, they may facilitate the spread of misinformation or deep fakes, negatively impacting society.

\end{document}